\documentclass[10pt,twocolumn,letterpaper]{article}

\usepackage{cvpr}      

\definecolor{cvprblue}{rgb}{0.21,0.49,0.74}
\usepackage[pagebackref,breaklinks,colorlinks,allcolors=cvprblue]{hyperref}

\usepackage[utf8]{inputenc}
\usepackage[T1]{fontenc}
\usepackage{graphicx}
\usepackage{booktabs}
\usepackage{multirow}
\usepackage{array}
\usepackage{makecell}
\usepackage{xcolor}
\usepackage{colortbl}
\usepackage{color}
\usepackage{hyperref}
\usepackage{CJKutf8}
\usepackage[most]{tcolorbox}
\usepackage{amsmath}
\usepackage{amsfonts}
\usepackage{nicefrac}
\usepackage{microtype}
\usepackage{multicol}
\usepackage{tikz}
\usepackage{wrapfig}
\usepackage{subcaption}
\usepackage{adjustbox}
\usepackage{tabularx}
\usepackage[capitalize]{cleveref}
\usepackage{longtable}
\usepackage{listings}
\usepackage{algorithm}
\usepackage{algpseudocode}
\usepackage{times}
\usepackage{latexsym}

\crefformat{section}{\S#2#1#3}

\hypersetup{
    colorlinks=true,
    linkcolor=black,
    citecolor=blue,
}

\title{Progressive Multimodal Reasoning via Active Retrieval}

\author{Guanting Dong\quad Chenghao Zhang\quad  Mengjie Deng\quad  Yutao Zhu\quad  Zhicheng Dou\thanks{\ \ Corresponding author.} \quad Ji-Rong Wen 
\\Gaoling School of Artificial Intelligence, Renmin University of China. \\
\texttt{\{dongguanting, dou\}@ruc.edu.cn} }

\begin{document}
\maketitle

\begin{abstract}

Multi-step multimodal reasoning tasks pose significant challenges for multimodal large language models (MLLMs), and finding effective ways to enhance their performance in such scenarios remains an unresolved issue. In this paper, we propose AR-MCTS, a universal framework designed to progressively improve the reasoning capabilities of MLLMs through Active Retrieval (AR) and Monte Carlo Tree Search (MCTS). Our approach begins with the development of a unified retrieval module that retrieves key supporting insights for solving complex reasoning problems from a hybrid-modal retrieval corpus.
 To bridge the gap in automated multimodal reasoning verification, we employ the MCTS algorithm combined with an active retrieval mechanism, which enables the automatic generation of step-wise annotations. This strategy dynamically retrieves key insights for each reasoning step, moving beyond traditional beam search sampling to improve the diversity and reliability of the reasoning space. Additionally, we introduce a process reward model that aligns progressively to support the automatic verification of multimodal reasoning tasks. Experimental results across three complex multimodal reasoning benchmarks confirm the effectiveness of the AR-MCTS framework in enhancing the performance of various multimodal models. Further analysis demonstrates that AR-MCTS can optimize sampling diversity and accuracy, yielding reliable multimodal reasoning. 
\end{abstract}

\section{Introduction}
\label{sec:intro}
Reasoning, as the fundamental capability of large language models (LLMs)~\citep{GPT-3.5, GPT-4, Llama-3, Qwen2} and multimodal large language models (MLLMs)~\citep{LLaVA, QWen-VL, MiniGPT-4, InternVL-1.5}, lays the foundation for generalization across a wide range of downstream tasks such as mathematical reasoning~\citep{wizardmath,scaling_math,MathVista} and visual question answering~\citep{vlm3,vqa,deepvqa}. In complex reasoning scenarios, models often require multiple steps to seek a final answer, with each reasoning step potentially generating several branches and resulting in various candidate reasoning paths. Therefore, efficiently identifying the correct path that includes key problem-solving steps while eliminating incorrect ones is essential. To achieve this, reasoning verification techniques~\citep{VerifyStep, RewardingProgress, math-shepherd} enable models to explore multiple candidate solutions and employ a high-quality reward model for path selection, thereby offering a promising approach to improve the reliability of model reasoning.

To improve the trustworthiness of complex reasoning, foundational efforts such as outcome reward models (ORMs)~\citep{ORM, DeepSeekMath} directly verify the quality of entire reasoning trajectories. However, ORMs can only provide sparse and result-oriented feedback. To obtain finer-grained verification, process reward models (PRMs)~\citep{VerifyStep, RewardingProgress, math-shepherd,Step-Aware-Verifier, pobf, rewardstep, Q*} are designed, offering intermediate rewards after each step and incorporating reinforcement learning from human feedback (RLHF) algorithms~\citep{RLHF} for backward supervision of generative models. Despite these advancements, the manual annotation of reasoning paths requires lots of human resources, limiting its scalability and applicability~\citep{beyondaccuracy}. In response to these challenges, recent developments in inference-time scaling~\citep{scalingtesttime, Beyond-Chinchilla-Optimal} have led to the integration of the MCTS algorithm into LLMs~\citep{mcts}. This combination allows models to autonomously sample potential reasoning paths at each step during the expansion process. Then, different value functions are designed to estimate the quality of each path, followed by back-propagation and pruning, finally achieving automatic step-level reasoning annotation without human effort~\citep{ppo-better, Automated-Process-Supervision, MCTSR, Preference-Trees, ReST-MCTS*, AlphaMath-Almost-Zero, BEATS, AlphaLLM, AlphaLLM-CPL, SC-MCTS*, Exlt}. 

While MCTS-based methods have been widely applied to text-based LLMs, their adaptation for MLLMs remains largely unexplored. Indeed, the distinct characteristics of multimodal scenarios require specialized adaptations of MCTS to effectively address their complexities. Let us illustrate the challenges by theoretically analyzing the limitations of existing MCTS-based methods. Given the input $x$ at each expansion step of MCTS, and the best reasoning path $y$ selected after simulation, the process could be modeled as:
\begin{align}
\label{eq:intro}
p(y \mid x) = \max_{i \in k} \ 
\underbrace{p_{\theta}(y \mid r_i, x)}_{\text{Simulation}} 
\cdot
\underbrace{p_{\phi}(r_i \mid x)}_{\text{Expansion}},
\end{align}
where $r$ and $k$ represent reasoning paths and the number of sampled paths, while $\phi$ and $\theta$ denote the generator and the verifier respectively. From this formulation, we can observe that both the expansion and simulation phases are crucial to the process, jointly determining the success of reasoning. Most existing approaches focus on optimizing the simulation process, while leaving the expansion process by beam search that relies on the model's internal knowledge~\citep{beam_limit2,beam_limit1}. This simple strategy is effective for text-only reasoning tasks, as LLMs are sufficiently pre-trained on text data and their internal knowledge can be accurately measured. However, in multimodal reasoning tasks, the internal knowledge of MLLMs is insufficient for reasoning path expansion, because the interactions between inputs from different modalities frequently encounter misalignment~\citep{mm_gap,misalign_mm}. Such errors will grow larger as each step in the reasoning process depends on the previous one, causing small mistakes to become bigger over time~\citep{self_check, dcot}. Consequently, developing effective strategies for reliable path expansion in MLLMs poses a significant challenge in multimodal reasoning tasks.

To address these problems, we propose to leverage retrieved external knowledge in reasoning path expansion, to enhance the path sampling quality in MCTS and improve MLLMs' capability in complex multimodal reasoning. Recent studies have confirmed that retrieval-augmented techniques can bridge knowledge gaps in multimodal reasoning~\citep{high-school-textbook,mrag-cot,rag_help_reasoning}, but they take all retrieved knowledge as a whole in the inference. Intuitively, each reasoning step requires different knowledge in a complex task. Therefore, we aim to dynamically provide the appropriate knowledge at \textit{each step} of the reasoning process, thereby enhancing the accuracy of reasoning paths. Furthermore, we propose to incorporate diverse problem-solving insights into both the expansion and value function of the MCTS algorithm. This integration is expected to not only expand the diversity of the sampling space but also enhance the reliability of the reasoning verification process.

Specifically, we propose \textbf{AR-MCTS}, a universal framework dedicated to progressively improving the complex reasoning capabilities of MLLMs through \textbf{A}ctive \textbf{R}etrieval and \textbf{M}onte \textbf{C}arlo \textbf{T}ree \textbf{S}earch. Specifically, we first design a unified retrieval module to retrieve key problem-solving insights for supporting complex reasoning from a hybrid-modal retrieval corpus. To further achieve reliable multimodal reasoning verification, we define the quality of each reasoning step as its potential to deduce the correct answer, enabling us to iteratively obtain step-wise annotations using the MCTS algorithm. Notably, we propose an active retrieval strategy during the MCTS expansion process, innovatively replacing beam search sampling with dynamically retrieved problem-solving insights, thereby enhancing both the diversity and reliability of the sampling space. Based on these fine-grained annotations, we progressively align a process reward model tailored for multimodal reasoning through step-wise Direct Preference Optimization (DPO)~\citep{dpo,step-dpo} and Supervised Fine-tuning (SFT) objectives, achieving automatic process-level reasoning verification.

Experimental results on three complex multimodal reasoning benchmarks demonstrate the effectiveness of AR-MCTS across various proprietary models. Further analysis reveals that AR-MCTS optimizes both sampling diversity and verification accuracy, providing a promising solution for reliable multimodal reasoning. 

In summary, our contributions are as follows:

\begin{itemize}[leftmargin=1em]

\item We theoretically model the key mechanisms of the MCTS-based approach in Equation~(\ref{eq:intro}), revealing its core limitations in solving multimodal reasoning problems.


\item We are the first to introduce the retrieval mechanism in each step of multimodal reasoning to replace traditional model self-sampling strategies, enhancing both sampling diversity and accuracy of multi-step reasoning.

\item We propose the AR-MCTS framework, which leverages the MCTS algorithm alongside an active retrieval strategy for improving multimodal reasoning. This framework automatically acquires high-quality step-wise reasoning annotations to progressively align a process reward model, ultimately enabling reliable automated multimodal reasoning verification.

\end{itemize}

\section{Related Work}
\label{sec:related}

\paragraph{LLM and MLLM Reasoning.}
Large Language Models (LLMs)~\cite{GPT-3.5,GPT-4,Llama-3,Qwen2,qwen2.5} and Multimodal Large Language Models (MLLMs)~\cite{LLaVA,QWen-VL,qwen2-vl,MiniGPT-4,internvl,InternVL-1.5,llava-onevision} have rapidly advanced, with broad applications in mathematics~\cite{Math-PUMA,math-shepherd}, programming~\cite{guo2024deepseekcoder,song2024cs}, medicine~\cite{li2024llavamed}, character recognition~\cite{wei2024general,qiao2024makingoracle}. Among their diverse capabilities, reasoning stands out as the most critical, serving as a foundational step toward universal understanding. Methods such as Chain-of-Thought (CoT)~\cite{cot}, Tree-of-Thought (ToT)~\cite{ToT}, and Program-of-Thought (PoT)~\cite{Pal,PoT} enhance logical coherence and response complexity by guiding models to decompose problems progressively, applying structured prompts and targeted training objectives, including multimodal tasks~\cite{mm-cot}. Another key approach is the reflection mechanism, which prompts models to iteratively evaluate and refine responses, leading to improved coherence. Prior studies~\cite{self-consistency} show that generating diverse reasoning paths and selecting the most consistent responses enhances LLM reasoning. 
Moreover, some efforts enhance the reasoning capability of LLM and MLLM by integrating data augmentation during the SFT phase~\citep{rft,wizardmath,mugglemath,shao2024deepseekmath}, along with utilizing external tools~\citep{tora,dotamath}. Research~\cite{math-shepherd, rewarding-progress,deepmind-solving-math,Q*,ORM,DeepSeekMath,Step-Aware-Verifier,rewardstep} has also highlighted robust reward models as a promising means to optimize response quality. Recently, OpenAI’s \textit{o1}~\footnote{\url{https://openai.com/index/openai-o1-system-card/}} introduced a “slow thinking” mechanism that, combined with Monte Carlo Tree Search (MCTS) strategies and verification models, simulates gradual reasoning to improve accuracy~\cite{MCTSR,SC-MCTS*,ReST-MCTS*}. While these advancements are focused on single-modal LLMs, the reasoning potential of MLLMs remains under explored.

\vspace{-4mm}
\paragraph{Multimodal Retrieval-Augmented Generation.}

 Recent Retrieval-Augmented Generation (RAG) has shown exceptional performance across various NLP tasks for LLMs by incorporating relevant information from diverse sources~\cite{DPA-RAG,self-rag,tan2024htmlraghtmlbetterplain,genirsuvery,dongkbqa,coral,flashrag,vifrag,ierc,chatkbqa}. This approach can also enhance reasoning and question-answering in the multimodal domain through cross-modal integration~\cite{RA-NLR,high-school-textbook,mrag-cot,mrag-embedding}. However, the reasoning process is largely unexplained and lacks verification mechanisms. In this paper, we propose an active retrieval strategy that retrieves multimodal information at each step to align the PRM, facilitating reliable reasoning verification.

\section{Preliminary}
\label{sec:preliminary}


\paragraph{Monte Carlo Tree Search.} MCTS is a widely used sampling-based search method for decision-making optimization. Its core algorithm consists of four steps: selection, expansion, evaluation, and back-propagation. By repeatedly executing these four steps, it constructs a search tree. During the selection phase, MCTS recursively selects child nodes from the root using the Upper Confidence Bound (UCB)~\citep{ucb}:
\begin{equation}
\small
\label{eq:ucb}
\text{UCB}(i)=w_i+C*\sqrt{2*\ln{\frac{N_i}{n_i}}},
\end{equation}


\vspace{-4mm}
\paragraph{Problem Formulation.} Formally, in multimodal reasoning, given a multimodal query $Q^m$ and corresponding retrieved problem-solving insights $r$ from the retrieved hybrid-modal corpus $D_{H}$, we assume that the MLLM \(\pi_\theta\) operates in an auto-regressive manner to generate a reasoning path of \(k\) steps \([y_1, \ldots, y_k]\):
\begin{equation}
\small
\label{eq:problem}
p_{\theta}(\boldsymbol{y} \mid Q^m, R)=\prod_{i=1}^{k} p_{\theta}\left(y_{i} \mid Q^m, r, y_{<i}\right).
\end{equation}
In this paper, we obtain different intermediate reasoning trajectories as the MLLM decodes to a specific termination token. Following the setup of ~\cite{tencent_self_improve}, we formulate the generation process as a Markov Decision Process (MDP)~\citep{mdp} and adopt sentence-level MCTS modeling. In reinforcement learning terminology~\citep{ppo}, we define the current decoded intermediate step \(y_i\) as a state \(s_i\), corresponding to a leaf node. The process of backtracking to sample the next step is considered an action \(a_i\). A  list of detailed definitions of MCTS for reasoning is given in the supplementary materials.

\begin{figure}[!t]
    \centering
    \includegraphics[width=0.95\linewidth]{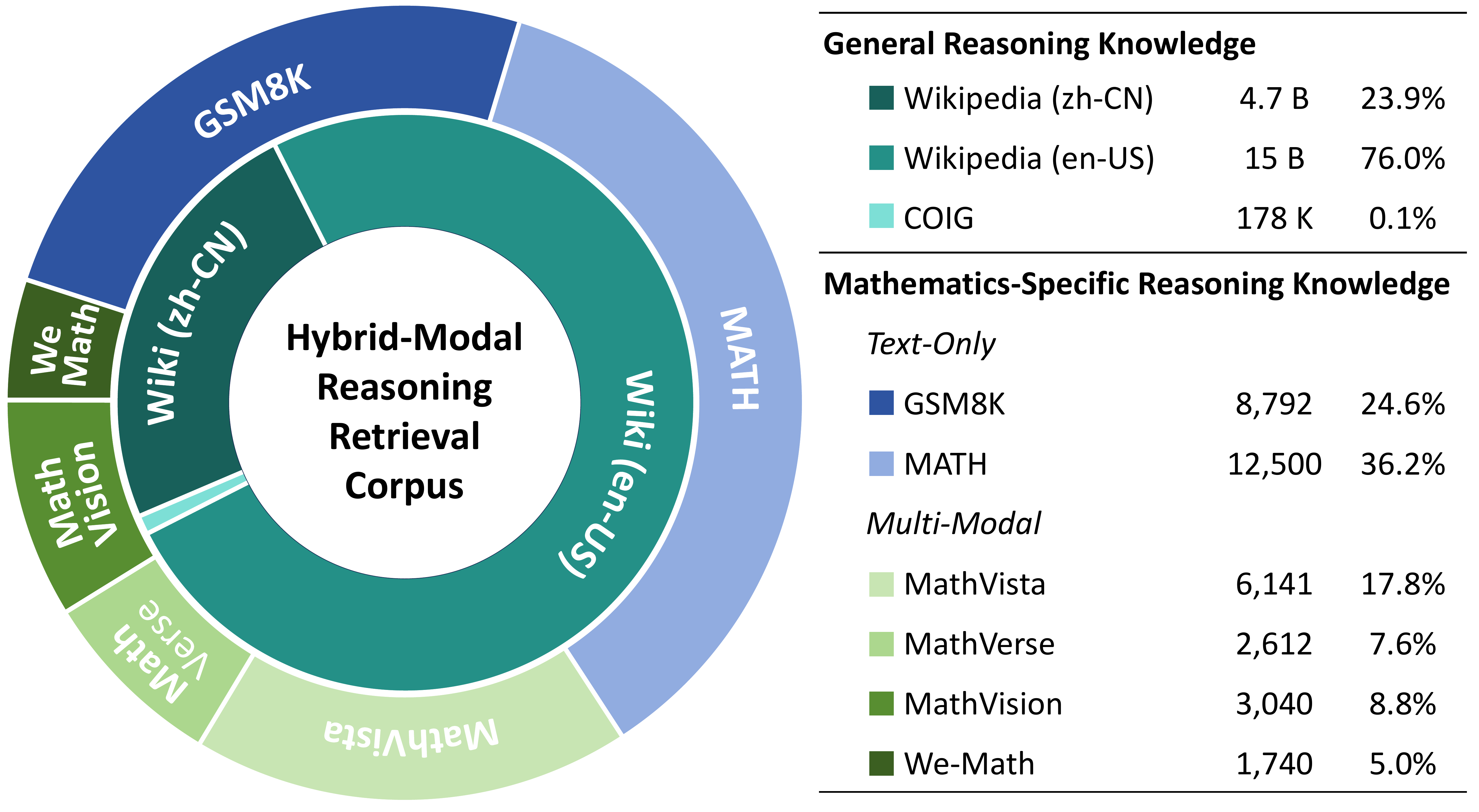}
     \vspace{-0.5em}
    \caption{The statistics of our hybrid-modal retrieval corpus.}
   
    \label{fig:retrieval_corpus}
    \vspace{-0.5em}
\end{figure}

\section{Methodology}


Our goal is to establish a process-level verification framework for improving multimodal reasoning without human annotation, while improving the diversity and accuarcy of candidate solution sampling. Therefore, we propose AR-MCTS framework to achieve fine-grained reasoning verification through active retrieval and Monte Carlo tree search. 

As shown in Figure \ref{fig:retrieval_corpus} \& \ref{fig:main}, AR-MCTS consists of two main components: 1) It introduces a unified retrieval module, including a high-quality hybrid-modal retrieval corpus (\Cref{sec:hybrid-modal retrieve corpus}) and a multimodal retrieval module (\Cref{sec:mmretrieval}). This module employs knowledge concept filtering to select key insights for problem-solving (\Cref{sec:kc_flitering}). 2) It automates the acquisition of step-wise annotations for multimodal reasoning using MCTS and an active retrieval mechanism (\Cref{sec:mmretrieval}). Then, it leverages the annotated data to progressively align the PRM in two stages (\Cref{sec:prm}), allowing for fine-grained verification of MLLM reasoning.







\subsection{Hybrid-Modal Retrieval Corpus Construction}
\label{sec:hybrid-modal retrieve corpus}

In an ideal scenario, improving reasoning capabilities through retrieval is akin to giving MLLMs an open-book exam. Unfortunately, the multimodal reasoning field consistently suffers from a lack of high-quality reasoning retrieval corpora. To systematically build a high-quality reasoning retrieval library, we conduct a comprehensive survey of open-source datasets, focusing on both general and mathematics-specific reasoning knowledge in multimodal reasoning.


\noindent\textbf{Mathematics-Specific Reasoning Knowledge.}\quad Mathematical reasoning is an essential skill of fundamental models, accompanied by the emergence of a series of high-quality datasets. In the text-only aspect, we select the most widely used mathematical reasoning datasets, GSM8K~\citep{gsm8k} and MATH~\citep{math}. For the multimodal domain, we adopt four meticulously cleaned high-quality multimodal math datasets: \textsc{MathVista}~\citep{MathVista}, MathVerse~\citep{MathVerse}, MathVision~\citep{MathVision}, and \textsc{We-Math}~\citep{WeMath}. To further prevent data leakage, we filtered out any overlapping portions with our testing benchmark using regular expressions, concatenating each sample's question $q$, solution process $p$, and answer $a$ into a single text format, along with the corresponding image storage paths. Ultimately, we obtain 22K text-only QA pairs and 12.5K multimodal sample pairs as proprietary sources \(D_{M}\) from six data sources, covering over 20 mathematical sub-fields, with each sample containing detailed solution steps.

\noindent\textbf{General Reasoning Knowledge.}\quad In the real world, general reasoning extends beyond natural subjects. To address this broader need, we follow the traditional RAG approach~\citep{traditional_rag, rag_survey} by utilizing the web-based retrieval source Wikipedia alongside the COIG~\citep{coig} large-scale question bank as our general reasoning retrieval sources. We conduct thorough data cleaning and chunking operations, ultimately constructing this extensive dataset as our general reasoning knowledge base \(D_{G}\). The statistical information of our hybrid-modal reasoning corpus $D_{\text{H}}=D_{M} \cup D_{G}$ is presented in Figure \ref{fig:retrieval_corpus}. More detailed information of processing retrieval corpus can be found in the supplementary.

\subsection{Unified Multimodal Retrieval Module}
\label{sec:mmretrieval}
Given a text-image pair from the multimodal test set $Q^{m} = \{x,t\}$, our goal is to retrieve the top-$K$ multimodal relevant knowledge for each sample. Since our retrieval library encompasses hybrid-modal retrieval sources, two retrieval processes are considered to obtain the top-K pairs:

\noindent\textbf{Text Retrieval.}\quad Given a text query \(q\) for multimodal sample, we aim to use a dense retriever to retrieve \(k\) relevant documents \(D_{q} = \{d_i\}_{i=1}^{k}\) from a text-only corpus. In this work, we employ Contriever~\citep{contriever} to obtain hidden vectors for both queries and documents. The relevance score is calculated by computing the dot-product similarity between the query and document representations, which facilitates the retrieval of the Top-\(K\) documents \(D_{q}\) as follow:
\begin{equation}
\small
D_{q} = \text{argtop}_k^{i = 1, \ldots, N} \left[ E_{\text{d}} (d_i)^{\top} \cdot E_{\text{q}}(q)  \right].
\end{equation}
\noindent\textbf{Cross-modal Retrieval.}\quad We utilize widely used contrastive vision-language models CLIP~\citep{CLIP}, which utilizes a dual-stream architecture featuring an image encoder \( E_{I}(\cdot) \) and a text encoder \( E_{T}(\cdot) \). Furthermore, we use CLIP to encode image-text pairs \((x, t)\), obtaining the image and text vectors \(E_{I}(x)\) and \(E_{T}(t)\). Since the hybrid-modal retrieval corpus contains both multimodal and text-only samples, we follow previous work~\citep{high-school-textbook} to derive encoding vectors for the entire hybrid-modal corpus $D_{\text{H}}$ as follows:
\begin{equation}
\label{eq:clip_encode}
\small
E_x (x,t) = \begin{cases}
\frac{E_{I}(x) + E_{T}(t)}{2}, & \text{if } t \neq \emptyset \text{ and } x \neq \emptyset, \\
E_{T}(t), & \text{if } t \neq \emptyset \text{ and } x = \emptyset.
\end{cases}
\end{equation}
where $\emptyset$ denotes empty set. For the \(i\)-th multimodal query \(Q^{m}\), we encode it into a mixed vector \(E_x(Q^{m}) = \frac{E_{I}(x) + E_{T}(t)}{2}\). We perform cross-modal retrieval between the encoding of each multimodal query and the entire retrieval database, utilizing FAISS~\citep{faiss-gpu} for indexing to retrieve \(K\) samples for each query:
\begin{equation}
\small
D_{\text{cross}} = \text{argtop}_k^{j = 1, \ldots, N} \left[ E_{x} (Q^{m})^{\top} \cdot E_{x} (x_j, t_j) \right].
\label{eq:clip_retrieve}
\end{equation}
Here, \(E_{x}(Q^{m})\) and \(E_{x}(x_j, t_j)\) denote the embeddings of the multimodal query and the samples in the hybrid-modal corpus, with indices \(j\) ranging from 1 to \(N\) to ensure that the entire retrieval database is considered.

\begin{figure}[!t]
    \centering
    \includegraphics[width=\linewidth]{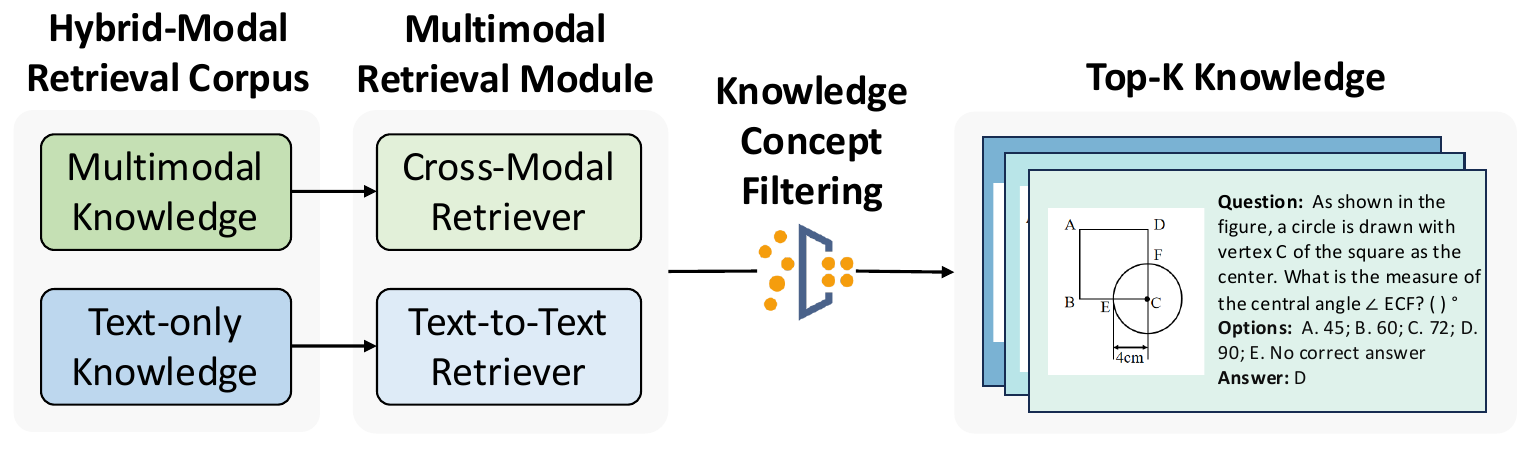}
     \vspace{-2em}
    \caption{The pipeline of our unified multimodal retrieval module.}
   
    \label{fig:retrieval_corpus}
    \vspace{-1em}
\end{figure}

\subsection{Knowledge Concept Filtering}
\label{sec:kc_flitering}
In our deployment process, we observe that multimodal reasoning with retrieved knowledge is highly sensitive to the consistency of fine-grained knowledge concepts (\textit{e.g.}, \textit{algebra} knowledge can't help in solving \textit{triangles} problem). Notably, most high-quality visual mathematical benchmarks provide detailed category labels (\textit{e.g.}, \textit{``Angles and Length''}) for each sample, motivating us to consider knowledge concept for fine-grained filtering. Given a multimodal query \(Q^m\) and its knowledge concept label \(L_{\text{kc}}\), we encode the top-\(K\) retrieved hybrid-modal samples from \(D_{\text{H}} = \{D_{\text{q}} \cup D_{\text{cross}}\}\) according to Equation~(\ref{eq:clip_encode}) and compute the similarity with the knowledge concept's embedding \(E_{T}(kc)\) following the pipeline in ``Cross-Modal Retrieval''. We strictly enforce the original retrieval similarity threshold \(T_r\) and the knowledge concept consistency threshold \(T_{\text{kc}}\), allowing only those samples that meet both criteria to serve as key insights \(D_{\text{ins}}\) for the query \(Q^m\):
\begin{equation}
\small
D_{\text{ins}} = \{r \in D_{\text{H}} \mid \text{Sim}(r, Q^m) \geq T_r \ \& \ \text{Sim}(r, L_{\text{kc}}) \geq T_{\text{kc}}\},\notag
\end{equation}
where $\text{Sim}(x,y)$ represents the cosine similarity between the embeddings $E(x)$ and $E(y)$, $r \in D_{\text{H}}$ denotes a retrieved insights from the corpus $D_{\text{H}}$. Detailed information of the filtering process can be found in supplementary materials.






\begin{figure*}[!t]
    \centering
    \includegraphics[width=0.88\linewidth]{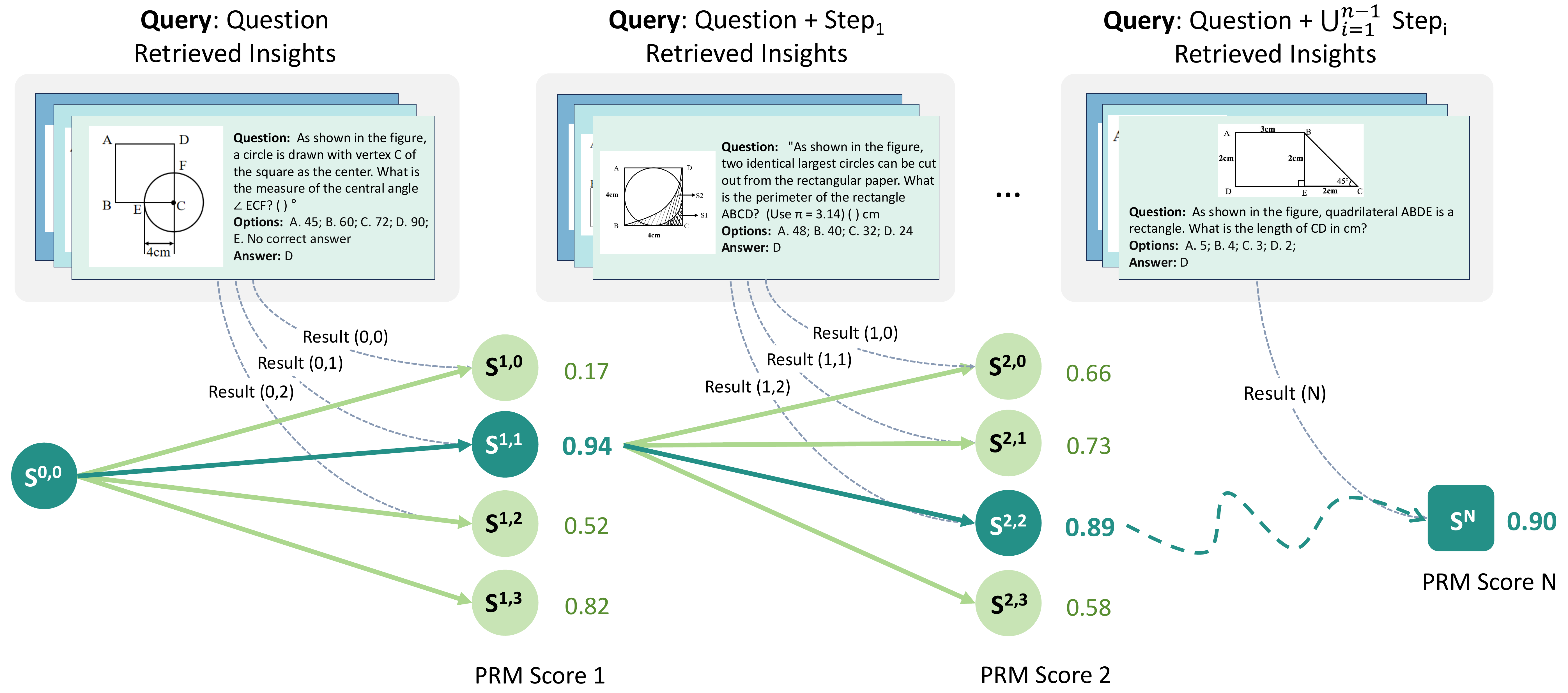}
    \caption{The overall framework of AR-MCTS: The retrieval module actively retrieves key insights at each step of the MCTS process. Then, the states of the MCTS is enhanced with different insights to expand the possible action space of the MLLM. Notably, one state of each step, such as state \(S^{1,3}\) and \(S^{2,3}\) in this figure, no insights are provided, and the state is a direct output of the MLLM.}
    \label{fig:main}
    \vspace{-0.5em}
\end{figure*}

\subsection{Progressive Multimodal Reasoning Annotation}
\label{sec:mcts}

In this section, we employ MCTS via active retrieval to facilitate MLLMs in the automatic generation of step-wise reasoning annotations, as shown in Figure \ref{fig:main}. Through the self-exploration process, we obtain \(Q\) values at each step (node) to capture potential reasoning errors in the intermediate steps. Below, we will present our detailed algorithm design, which includes four core operations:


\begin{itemize}[leftmargin=1em]

\item \textbf{Selection.} During the $j$-th simulation of the AR-MCTS, the process begins with $s_0$, representing the initial state containing the multimodal input query $Q_0^m = (x_0, t_0)$ and corresponding retrieved problem-solving insights $r_0$. The algorithm then proceeds to explore the Monte Carlo tree by selecting as Equation~(\ref{eq:ucb}) iteratively, then we can formulate the multimodal query of state $s_j$ as $Q^m_j = \{(x_j, t_j) \mid t_j = t_0+ \sum_{i=1}^{j} y_i\}$, as shown in Figure \ref{fig:main}.

\vspace{2mm}
\item \textbf{Expansion with Active Retrieval Strategy.} Given the state \(s_i\) represented by the selected leaf node, the MCTS-based approach backtracks to the prior state, forming our multimodal input as $(x_i,t_i, r_i)$. The temperature in the traditional expansion process is empirically increased to greater than 0.6 to sample multiple potential candidate actions for the next step~\citep{alibaba_step_level}. Unlike them, we emphasize that the supporting knowledge required for different reasoning trajectory at each step should vary, and propose an \textbf{Active Retrieval} strategy. As shown in Figure \ref{fig:main}, during the MCTS expansion phase at state \(s_i\), we first concatenate the input \(Q^m_i\) with the previous reasoning steps. Then we dynamically retrieve the required candidate insights \(r_i\) for each step from the problem-solving insight library \(D_{\text{ins}}\) according to Equation~(\ref{eq:clip_retrieve}), and replace the insight \(r_{i-1}\) from the previous step with the latest retrieved insights \(r_i\). According to the Equation~(\ref{eq:problem}), the process of sampling \(k\) reasoning paths at state \(s_i\) can be modeled as follows:
\begin{equation}
\small
p_{\theta} (y \mid x)=\prod_{i=1}^{k} p_{\theta}\left(\{y_{i}^j\}_{j=1}^k \mid Q_i^m, r_{i}\right).
\end{equation}
\item \textbf{Simulation.}
We use the probability of deducing the correct answer based on partial solutions as a criterion for quality assessment. Following \citeauthor{math-shepherd}, we apply a one-step rollout for each node obtained during expansion to ensure efficiency, and we construct a value function as $V(s_i) = \frac{\sum_{j=1}^{k} \mathbb{I}(y_{j} = \hat{y_{i}})}{k}$, where \(k\), \(\mathbb{I}\) denotes the number of sampled reasoning paths and the indicator function. If the final answer \(y_{j}\) equals the grounding truth \(\hat{y_{i}}\), we set the value of the current node to 1; Otherwise, we set it to 0.


\vspace{2mm}
\item \textbf{Back-Propagation.}
For the terminal nodes reached during the rollout and the current leaf node, MCTS performs a backward update of the visit count and Q-value for each \((s, a)\) along the route from the current node to the root, which is fomulated as $N(s, a) \leftarrow N(s, a) + 1$, $ Q(s, a) \leftarrow Q(s, a) + \frac{1}{N(s, a)} \left(V(s) - Q(s, a)\right)$.
\end{itemize}

\subsection{Curriculum Process Reward Modeling}
\label{sec:prm}


After acquiring step-wise reasoning annotations, we draw inspiration from curriculum learning~\citep{easy2hard, how_abilities} to design a two-phase approach for PRM. In the first stage, the model learns to distinguish the correctness of reasoning steps. In the second stage, it learns to assign scores to each step, facilitating generalization from easy to hard.


\noindent\textbf{Step-wise DPO Pre-alignment.}\quad In the first phrase, each round of expansion and evaluation in AR-MCTS naturally generates batches of positive and negative pairs, inspiring us to align preferences using step-level Direct Preference Optimization (DPO) as the training objective. Under the state \( s_i \) ($i$-th step in reasoning), given a multimodal query \( Q_i^m = (x_i, t_i) \) and a sampled reasoning paths set $Y_i = \{y_j\}_{j=1}^k$ , along with the corresponding value set \( V_i = \{v_j\}_{j=1}^k \). We filter the solution paths in $Y_i$ with value \(v_j > 0.8\) as positive samples \(y_j^+\), while those with \(v_j = 0\) are regarded as negative samples \(y_j^-\). Therefore, for each problem $Q_i^m$, we can obtain $K$ pairs of step-level preference pairs $D_i^{\text{step}} = (y_j^+, y_j^-)_{j=1}^K$ and follow step-level DPO to align the reasoning discernment capability as:
\begin{gather*}
\mathcal{L}_{\text{SDPO}}(\pi_\theta;\pi_\text{ref}) = - \mathbb{E}_{(Q^m, y^{+}, y^{-}) \sim \mathcal{D}^{\text{step}}} [\text{log}\sigma (  \beta  \text{log} \\ \frac{\pi_\theta(y^+|Q^m)}{\pi_\theta(y^+|Q^m)} 
- \beta \text{log}\frac{\pi_\text{ref}(y^-|Q^m)}{\pi_\text{ref}(y^-|Q^m)})],
\end{gather*}



The reference model \( \pi_\text{ref} \) is initially set to \( \pi_\theta \) and remains constant during training. Here, \( \beta \) is a hyperparameter, and \( \sigma \) denotes the sigmoid function. The objective of \( \mathcal{L}_{\text{SDPO}} \) is to maximize the likelihood of preferred \( y^+ \) compared to the dispreferred \( y^- \).

\noindent\textbf{Point-wise Fine-tuning.}\quad 
After pre-alignment, our PRM has gained the initial ability to distinguish the correctness of step-wise reasoning. To further unlock its reasoning scoring capability, we apply a step-level cross-entropy objective to the pre-aligned PRM \(\pi_\text{DPO}\) using the following parameters:
\begin{equation}
\label{eq:sft}
\small
\mathcal{L}_{\text{PFT}} = \sum_{i=1}^{N} \ \left[\hat{y_{i}}\log_{\pi_\text{SDPO}} \left(r_{i}\right) + (1 - \hat{y_{i}})\log_{\pi_\text{SDPO}}(1 - r_i) \right], \notag
\end{equation}
where \( \hat{y_{i}} \) is the golden label $(0,1)$ for the state $s_i$, \( r_{i} \) is the sigmoid score assigned by PRM. After the above two stages, we progressively achieve an aligned process reward model.



\noindent\textbf{Inference.}\quad During the inference phase of AR-MCTS, we utilize the fine-tuned PRM to follow the steps of AR-MCTS in Figure \ref{fig:main}, employing the PRM scores as the value for each step in the evaluation phrase. Following \citeauthor{google_PRM}, we adopt point-wise soft labels. We also make a discussion of PRM’s hard labels in the supplementary materials. Unlike the annotation process for training data, we extract the top-scoring node from the K expansion reasoning paths each round, discarding the other low-quality paths. Moreover, we set an early stopping criterion of 4, which allows us to derive the final result directly in the 4-th round to reduce computational complexity.

\begin{table*}[t] 
    \centering 
    \footnotesize
    \caption{Mathematical reasoning assessment on different MLLMs using \textsc{MathVista} and \textsc{We-Math} \textit{testmini} Sets. In the case of \textsc{MathVista}, we picked 6 categories from the original 12: ALL (overall accuracy), GPS (geometry problem solving), MWP (math word problems), ALG (algebraic reasoning), GEO (geometry reasoning), and STA (statistical reasoning). For \textsc{We-Math}, we selected 8 categories: S1 (one-step problems), S2 (two-step problems), S3 (three-step problems), AVG (strict overall average scores), IK (insufficient knowledge), IG (inadequate generalization), CM (complete mastery), and RM (rote memorization). The top scores for each model are highlighted in \textbf{bold}.}
    \vspace{-0.5em}
    \setlength{\tabcolsep}{3.3pt}{
        \begin{tabular}{llcccccccccccccc}
            \toprule
            \multirow{2}{*}[-2pt]{Model} & \multirow{2}{*}[-2pt]{Method} & \multicolumn{6}{c}{\textsc{\textbf{MathVista}}} & \multicolumn{8}{c}{\textsc{\textbf{We-Math}}} \\
            \cmidrule(r){3-8} \cmidrule(l){9-16}
            & & ALL $\uparrow$ & GPS $\uparrow$ & MWP $\uparrow$ & ALG $\uparrow$ & GEO $\uparrow$ & STA $\uparrow$ & S1 $\uparrow$ & S2 $\uparrow$ & S3 $\uparrow$ & AVG $\uparrow$ & IK $\uparrow$ & IG $\uparrow$ & CM $\uparrow$ & RM $\downarrow$ \\
            \cmidrule{1-16}
            \multirow{5}*{GPT-4o}
            & Zero-shot & 59.0 & 59.6 & 65.1 & 61.2 & 60.7 & 72.4 & 71.5 & 58.3 & 46.1 & 40.8 & \textbf{31.8} & 13.7 & 33.9 & 37.8 \\
            & Self-Consistency & 61.8 & 68.3 & 65.1 & \textbf{68.0} & 68.2 & 74.8 & 73.3 & 63.6 & 53.0 & 45.2 & 29.9 & 12.8 & 38.8 & 32.8 \\
            & Self-Correction & 59.9 & 61.1 & 65.6 & 61.2 & 61.1 & 72.8 & 72.8 & 58.9 & 43.6 & 42.9 & 31.2 & \textbf{15.2} & 35.2 & 34.2 \\
            & ORM & 61.9 & 68.3 & 66.1 & \textbf{68.0} & 68.2 & 74.8 & 73.1 & 63.3 & 50.3 & 44.3 & 26.5 & 10.9 & 38.9 & 38.0 \\
            & \cellcolor{gray!15}AR-MCTS & \cellcolor{gray!15}\textbf{62.6} & \cellcolor{gray!15}\textbf{68.6} & \cellcolor{gray!15}\textbf{66.4} & \cellcolor{gray!15}\textbf{68.0} & \cellcolor{gray!15}\textbf{68.8} & \cellcolor{gray!15}\textbf{75.3} & \cellcolor{gray!15}\textbf{74.7} & \cellcolor{gray!15}\textbf{65.6} & \cellcolor{gray!15}\textbf{56.4} & \cellcolor{gray!15}\textbf{46.8} & \cellcolor{gray!15}28.0 & \cellcolor{gray!15}12.8 & \cellcolor{gray!15}\textbf{40.4} & \cellcolor{gray!15}\textbf{31.8} \\
            \midrule
            \multirow{5}{2.1cm}{LLaVA-\\OneVision-72B} 
            & Zero-shot & 64.2 & \textbf{80.8} & 69.4 & 73.3 & \textbf{77.0} & 66.8 & 58.1 & 44.7 & 40.6 & 24.6 & 42.5 & 14.1 & 17.5 & 59.7 \\
            & Self-Consistency & 66.0 & 79.8 & \textbf{73.1} & 74.0 & 76.6 & \textbf{67.8} & 70.7 & \textbf{52.8} & 38.2 & 36.9 & 33.9 & 15.8 & \textbf{29.0} & 42.4 \\
            & Self-Correction & 58.3 & 78.4 & 68.8 & 70.1 & 74.9 & 56.8 & 48.2 & 33.9 & 30.3 & 14.7 & \textbf{55.4} & 11.8 & 8.7 & 73.3 \\
            & ORM & 65.9 & 80.3 & \textbf{73.1} & 74.0 & \textbf{77.0} & \textbf{67.8} & 66.6 & 48.3 & \textbf{44.2} & 30.6 & 34.9 & \textbf{18.1} & 21.5 & 54.3 \\
            & \cellcolor{gray!15}AR-MCTS & \cellcolor{gray!15}\textbf{66.3} & \cellcolor{gray!15}79.8 & \cellcolor{gray!15}\textbf{73.1} & \cellcolor{gray!15}\textbf{74.4} & \cellcolor{gray!15}76.6 & \cellcolor{gray!15}\textbf{67.8} & \cellcolor{gray!15}\textbf{71.1} & \cellcolor{gray!15}\textbf{52.8} & \cellcolor{gray!15}38.9 & \cellcolor{gray!15}\textbf{37.4} & \cellcolor{gray!15}33.7 & \cellcolor{gray!15}\textbf{18.1} & \cellcolor{gray!15}28.4 & \cellcolor{gray!15}\textbf{41.1} \\
            \midrule
            \multirow{5}{*}{InternVL2-8B} 
            & Zero-shot & 57.3 & 62.5 & 62.4 & 61.2 & 60.7 & 59.1 & 50.0 & 36.7 & 23.6 & 17.4 & 59.8 & 10.1 & 12.4 & 58.9 \\
            & Self-Consistency & 61.8 & \textbf{77.4} & 64.0 & \textbf{73.0} & \textbf{72.8} & 62.1 & 58.4 & 47.1 & 35.1 & 26.6 & 45.5 & 13.5 & 19.8 & 51.6 \\
            & Self-Correction & 46.8 & 57.7 & 31.2 & 55.9 & 56.1 & 46.2 & 43.5 & 28.1 & 30.3 & 9.8 & \textbf{62.7} & 8.6 & 5.5 & 80.8 \\
            & ORM & 61.1 & 67.8 & 64.0 & 64.1 & 64.9 & 68.4 & 64.0 & 45.0 & 32.7 & 29.7 & 42.9 & \textbf{16.0} & 21.7 & \textbf{47.2} \\
            & \cellcolor{gray!15}AR-MCTS & \cellcolor{gray!15}\textbf{63.1} & \cellcolor{gray!15}62.9 & \cellcolor{gray!15}\textbf{71.6} & \cellcolor{gray!15}59.9 & \cellcolor{gray!15}62.6 & \cellcolor{gray!15}\textbf{71.4} & \cellcolor{gray!15}\textbf{65.1} & \cellcolor{gray!15}\textbf{52.2} & \cellcolor{gray!15}\textbf{43.6} & \cellcolor{gray!15}\textbf{30.5} & \cellcolor{gray!15}37.7 & \cellcolor{gray!15}14.7 & \cellcolor{gray!15}\textbf{23.2} & \cellcolor{gray!15}51.2 \\
            \midrule
            \multirow{5}{*}{Qwen2-VL-7B} 
            & Zero-shot & 58.8 & 45.5 & 60.5 & 45.5 & 47.9 & 70.8 & 53.4 & 37.2 & 33.9 & 19.8 & 51.2 & 12.6 & 13.5 & 62.6 \\
            & Self-Consistency & 61.2 & 54.8 & 61.8 & 56.2 & 55.2 & 72.1 & 57.6 & 41.9 & 33.9 & 23.6 & 46.9 & 13.7 & 16.8 & 57.5 \\
            & Self-Correction & 50.8 & 43.3 & 53.2 & 45.9 & 43.9 & 62.1 & 52.3 & 38.6 & 26.7 & 20.0 & \textbf{54.1} & 11.1 & 14.5 & 58.5 \\
            & ORM & 62.3 & 55.5 & 62.7 & 56.9 & 56.5 & \textbf{72.4} & 57.8 & 45.1 & 34.6 & 26.4 & 42.9 & 11.2 & 20.8 & 54.8 \\
            & \cellcolor{gray!15}AR-MCTS & \cellcolor{gray!15}\textbf{64.1} & \cellcolor{gray!15}\textbf{63.9} & \cellcolor{gray!15}\textbf{72.6} & \cellcolor{gray!15}\textbf{60.9} & \cellcolor{gray!15}\textbf{63.6} & \cellcolor{gray!15}\textbf{72.4} & \cellcolor{gray!15}\textbf{59.9} & \cellcolor{gray!15}\textbf{48.1} & \cellcolor{gray!15}\textbf{40.6} & \cellcolor{gray!15}\textbf{28.1} & \cellcolor{gray!15}40.0 & \cellcolor{gray!15}\textbf{14.3} & \cellcolor{gray!15}\textbf{21.0} & \cellcolor{gray!15}\textbf{54.2} \\
            \bottomrule
        \end{tabular}
    }
    \label{tab:main_result}
\end{table*}

\section{Experiments}

\subsection{Experimental Setup}

To assess the effectiveness of the AR-MCTS, we provide a detailed introduction from the following aspects:

\vspace{-4mm}
\paragraph{Benchmarks and Baselines.} We perform experiments on two widely used multimodal mathematical reasoning benchmarks: \textsc{MathVista}~\citep{MathVista} and \textsc{We-Math}~\citep{WeMath}. To further validate our AR-MCTS in general reasoning domain, we perform cross-domain evaluation on the GAOKAO-MM benchmark~\citep{gaokaomm}. For baselines, we employ AR-MCTS on strong proprietary and open-source models: (1) Closed-source MLLMs: GPT-4o~\citep{openai2024gpt4ocard}, GPT-4V~\citep{2023GPT4VisionSC}; (2) Open-source MLLMs: LLaVA-OneVision-Qwen2-72B~\citep{llava_onevision}, InternVL2-8B~\citep{internvl2}, Qwen2-VL-7B~\citep{qwen2vl}, LLaMA3-LlaVA-NeXT-8B~\citep{liu2024llavanext}. Referencing relevant works on MCTS~\citep{math-shepherd,tencent_self_improve}, we implement Self-Consistency~\citep{self-consistency}, Self-Correction~\citep{Self-Correction}, and ORM~\citep{ORM} as our core comparison strategies.



 \begin{table*}[t] 
    \centering 
    \footnotesize
    \caption{The Performance of MLLMs on GAOKAO-MM. The top scores for each model are highlighted in \textbf{bold}.}
    \vspace{-0.5em}
    \setlength{\tabcolsep}{6pt}{
        \begin{tabular}{llccccccccc}
			\toprule
			Model & Method & Overall & Mathematics & Chinese & Physics & Chemistry & Biology & History & Geography & Politics \\
            \cmidrule{1-11}
            \multirow{3}*{GPT-4o}
            & Zero-shot & 45.6 & 50.0 & 33.0 & 9.6 & 35.7 & \textbf{50.0} & 60.0 & \textbf{73.1} & \textbf{100.0} \\
            & Self-Consistency & 47.8 & 50.0 & 33.0 & 13.5 & \textbf{42.9} & \textbf{50.0} & 60.0 & \textbf{73.1} & \textbf{100.0} \\
            & \cellcolor{gray!15}AR-MCTS & \cellcolor{gray!15}\textbf{52.2} & \cellcolor{gray!15}\textbf{62.5} & \cellcolor{gray!15}\textbf{33.3} & \cellcolor{gray!15}\textbf{21.2} & \cellcolor{gray!15}\textbf{42.9} & \cellcolor{gray!15}\textbf{50.0} & \cellcolor{gray!15}\textbf{80.0} & \cellcolor{gray!15}\textbf{73.1} & \cellcolor{gray!15}\textbf{100.0}\\
            
            \midrule
            \multirow{3}*{Qwen2-VL-7B}
            & Zero-shot & 30.2 & 25.0 & \textbf{33.3} & \textbf{21.2} & 42.9 & \textbf{50.0} & \textbf{40.0} & 26.9 & 40.0 \\
            & Self-Consistency & 33.0 & \textbf{50.0} & 33.0 & 15.4 & \textbf{50.0} & 25.0 & 20.0 & 38.5 & 40.0 \\
            & \cellcolor{gray!15}AR-MCTS & \cellcolor{gray!15}\textbf{37.4} & \cellcolor{gray!15}37.5 & \cellcolor{gray!15}\textbf{33.3} & \cellcolor{gray!15}19.2 & \cellcolor{gray!15}35.7 & \cellcolor{gray!15}\textbf{50.0} & \cellcolor{gray!15}\textbf{40.0} & \cellcolor{gray!15}\textbf{46.2} & \cellcolor{gray!15}\textbf{80.0}\\

			\bottomrule
		\end{tabular}
 	}
    \vspace{-0.5em}
	\label{tab:gaokao_result}
\end{table*}

\vspace{-4mm}
\paragraph{Data Sampling via AR-MCTS.}
As highlighted by MathPUMA~\citep{Math-PUMA}, the challenge arises because the three multimodal benchmarks we evaluate lack training sets. Following the collection described in \Cref{sec:hybrid-modal retrieve corpus}, we utilize four multimodal and two text-only datasets for process annotation, excluding any sources currently under evaluation. We extract multimodal QA pairs and use our AR-MCTS algorithm to automatically generate and annotate detailed solution processes. Notably, the GAOKAO-MM dataset is entirely in Chinese, which complicates reliance on English data sources. To address this, we classify data from 2010 to 2021 for AR-MCTS annotation, while questions from 2022 to 2023 serve as the test set. For a comprehensive overview of the experimental setup, please find in the supplementary materials..

\subsection{Overall Results}
Table~\ref{tab:main_result} illustrates the main results. Overall, AR-MCTS significantly improves multimodal reasoning performance across various MLLMs and reasoning verification strategies (Self-Correction, Self-Consistency) on two benchmarks, conclusively demonstrating the advantages of our approach. Furthermore, we have identified the following insights:

\noindent\textbf{MLLMs struggle to self-correct reasoning errors.}\quad The self-correction strategy struggles across both reasoning benchmarks. Although a minor improvement is noted with GPT-4o, other weaker open-source MLLMs experience significant declines after the self-correction process, particularly Qwen2VL-7B, which shows a drop of over 8\% on \textsc{MathVista} (ALL). This discovery corresponds with the findings of \citeauthor{Self-Correction}, highlighting the instability of correction methods that rely on the self-knowledge of MLLMs in multimodal reasoning, especially in models with fewer parameters.

\noindent\textbf{PRM outperforms ORM in complex reasoning tasks.}\quad \textsc{We-Math} is a step-level evaluation featuring S1 to S3, which progressively increases the difficulty of reasoning steps. Compared to ORM, AR-MCTS with PRM demonstrates a more significant performance improvement across most MLLM backbones on the S3 metrics in \textsc{We-Math} (GPT-4o: 56.4\% vs 50.3\%; Qwen2-VL: 40.6\% vs 34.6 \%). This highlights that PRM, by meticulously verifying each step of the reasoning process, achieves stronger alignment in multi-step reasoning tasks.

\noindent\textbf{AR-MCTS better unlocks the reasoning potential of weaker MLLMs.}\quad Compared to LLaVA-OneVision-72B, Qwen2-VL-7B with AR-MCTS shows a significant improvement over the zero-shot setting on \textsc{MathVista} (ALL: 5.3\%$\uparrow$) and in \textsc{We-Math} (AVG: 8.3\%$\uparrow$). A similar conclusion is observed with InternVL2-8B, indicating that the performance gains of AR-MCTS are more pronounced in smaller MLLMs. To gain insight into this result in conjunction with Equation~(\ref{eq:intro}), we consider that, under the same verifier, weaker MLLMs may sample the correct answers but struggle to directly decode those paths greedily. This suggests that smaller MLLMs have the potential for correct reasoning but may not successfully decode answers relying solely on internal knowledge. This observation further verifies the importance of integrating active retrieval in multimodal reasoning. It also demonstrates that AR-MCTS is a reliable and plug-and-play framework, offering a promising solution for reasoning alignment in weaker MLLMs.

   

\subsection{General Reasoning Domain Verification}
To validate the effectiveness of AR-MCTS in the general multimodal reasoning field, we further evaluate the Chinese human-level multimodal reasoning benchmark, GAOKAO-MM. As shown in Table \ref{tab:gaokao_result}, both the closed-source model GPT-4o and the open-source small model Qwen2-VL-7B demonstrate significant improvements over the backbone and self-consistency approaches when combined with the AR-MCTS framework, verifying the generalization of AR-MCTS across different languages and reasoning disciplines. Notably, AR-MCTS with GPT-4o achieves stable improvements in mathematics and physics (12.5\% $\uparrow$ and 7.7\% $\uparrow$), while also showing some gains in the humanities (e.g. history 20\%$\uparrow$). This emphasizes that AR-MCTS with PRM not only improves the complex reasoning capabilities of MLLMs, but also effectively mitigates the knowledge gaps of MLLMs in the humanities through its active retrieval mechanism.


\begin{table}[t!] 
\centering 
\footnotesize
    \caption{Ablation study with Qwen2-7B. "Filtering" denotes the knowledge concept filtering module.}
    \vspace{-1em}
    \begin{tabular}{lccc}
        \toprule
        {Models} & {\makecell[c]{\textsc{MathVista}\\(ALL)}} & {\makecell*[c]{\textsc{We-Math}\\(S3)}} & {\makecell*[c]{GAOKAO\\-MM(ALL)}} \\ 
        \midrule
        AR-MCTS & 64.1 & 40.6 & 37.4 \\
        \quad \textit{w/o} PRM & 61.0 {\scriptsize (-3.1)} & 37.7 {\scriptsize (-2.9)} & 33.2 {\scriptsize (-4.2)} \\
        \quad \textit{w/o} Filtering & 62.8 {\scriptsize(-1.3)} & 39.5 {\scriptsize(-1.1)} & 34.5 {\scriptsize (-2.9)} \\
        \quad \textit{w/o} Active Retrieval & 61.9 {\scriptsize(-2.2)} & 38.7 {\scriptsize(-1.9)} & 33.4 {\scriptsize(-4.0)} \\
        \bottomrule
    \end{tabular}
    \label{tab:ablation_main}
\end{table}


\subsection{Quantitative Analysis}
\noindent\textbf{Ablation Study.}\quad To explore the effects of various components in AR-MCTS, we conduct an ablation study in Table \ref{tab:ablation_main}. The term "w/o" indicates versions where specific components are removed. Our key observations are:
1) Removing any component from AR-MCTS results in performance decline, highlighting the necessity of all component designs.
2) Removing the PRM and active retrieval mechanism leads to a significant performance drop respectively (\textsc{MathVista}: 3.1\% \& 2.2\%), demonstrating that step-wise verification and active retrieved knowledge can effectively improve multimodal reasoning capabilities.
3) Notably, knowledge concept filtering also achieves stable performance gains, indicating that it effectively reduces noise in retrieved knowledge and highlights the critical importance of consistency between the retrieved knowledge and the problem during reasoning. Detailed ablations can be found in the supplementary materials.

\begin{figure}[t]
    \centering
    \resizebox{\columnwidth}{!}{
    \includegraphics{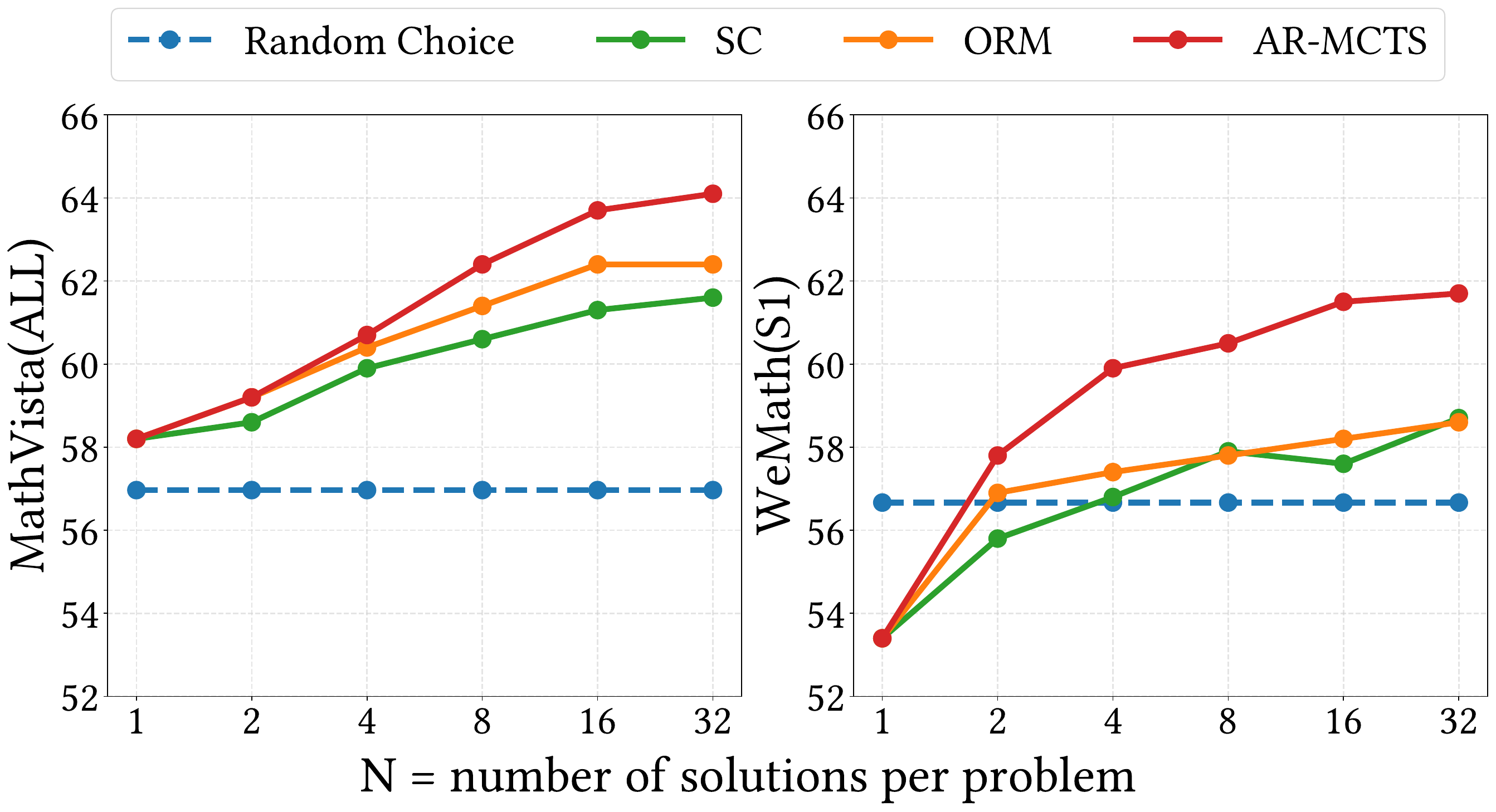}
    }
    \vspace{-1em}
    \caption{Scaling analysis on inference samplings. \textit{Random Choice} denotes the average result of randomly sampling from 1 to 32. }
    \label{fig:scaling}
\end{figure}


\noindent\textbf{Scaling Analysis on Inference Samplings.}\quad In this section, we conduct a scaling analysis to evaluate the performance of different strategies across two benchmarks with varying numbers of candidate solution paths, ranging from 1 to 32. As shown in Figure \ref{fig:scaling}, our main observations are as follows: 1) When the number of candidate solutions exceeds a certain threshold (16), self-consistency (SC) exhibits some performance fluctuations in \textsc{We-Math}. 2) AC-MCTS consistently outperforms ORM and SC, with this superiority becoming more pronounced as N increases. We attribute this advantage to our automated process labeling, which offers high scalability and low annotation costs while providing more reliable feedback for the verification of each path.

\subsection{Does AR-MCTS Improve the Sampling Space?}

\begin{figure}[t]
    \centering
    \resizebox{\columnwidth}{!}{ 
    \includegraphics{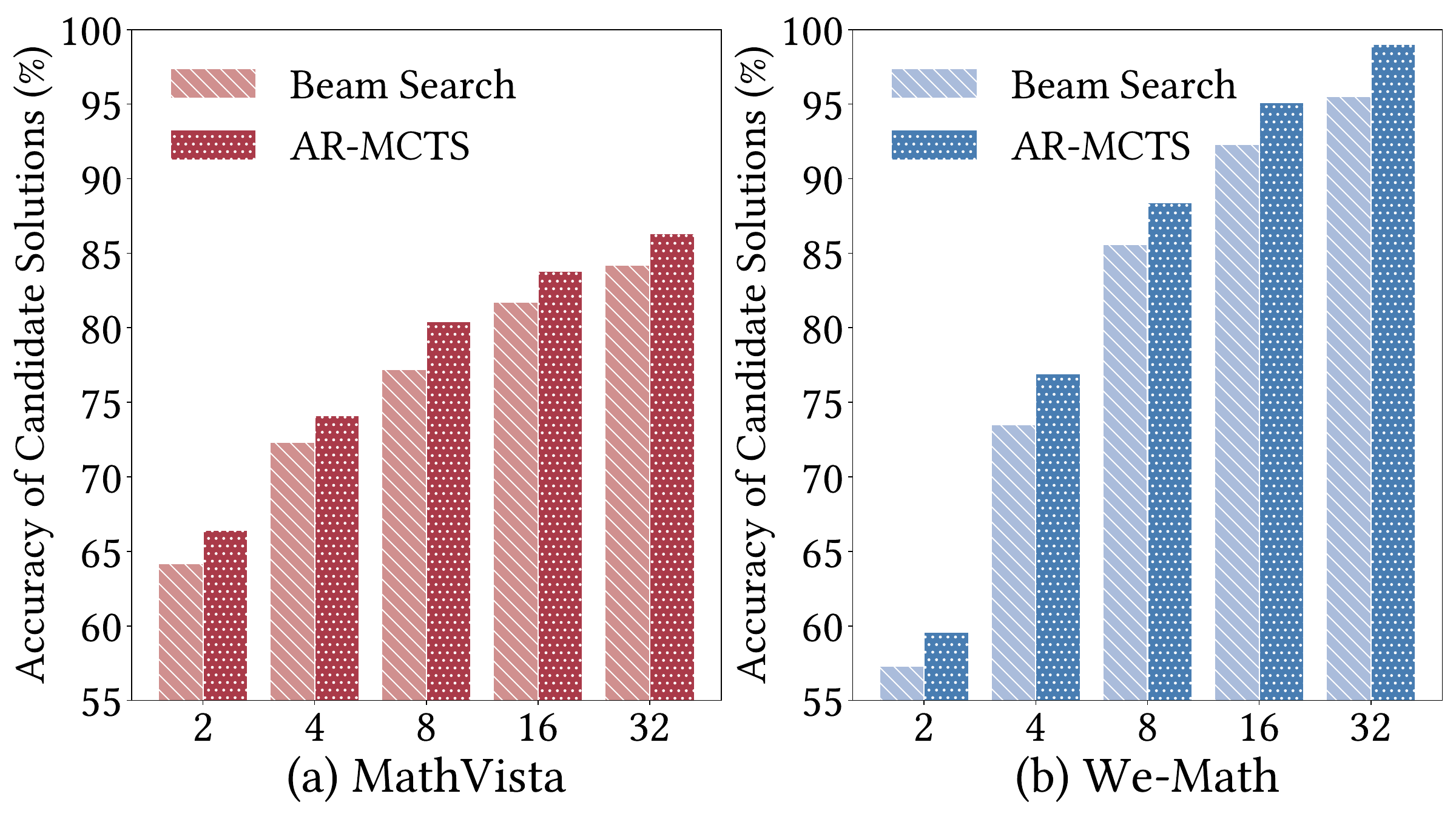}
    }
    \vspace{-2em}
    \caption{The accuracy comparison of solution sampling.}
    \label{fig:acc_retrieve}
    \vspace{-0.5em}
\end{figure}

In this section, we explore that AR-MCTS can efficiently improve the quality of the candidate solution sampling from the following two perspectives:

\noindent\textbf{Accuracy Analysis.}\quad To validate that AR-MCTS can efficiently improve the solution sampling accuracy in multimodal reasoning, we perform a quantitative analysis of the "Correctness of questions" during the sampling process of Qwen2-VL in the \textsc{MathVista} and \textsc{We-Math}. The accuracy can be formulated as  $P_{Q}^{\text{c}} = \frac{N_{Q}^{\text{c}}}{N_{Q}}$, where $N_{Q}^{\text{c}}$ denote the number of questions contain at least one correct candidate solution, while $N_{Q}$ denotes all number of questions.
As shown in Figure \ref{fig:acc_retrieve}, AR-MCTS demonstrates consistent gains in both benchmarks compared to the traditional beam search sampling. Moreover, as the number of candidate solutions increases, the answer accuracy $P_{Q}^{\text{c}}$ exhibits a positive correlation. This finding further confirms that AR-MCTS can efficiently improve the reliability of the sampling space in multimodal reasoning, effectively addressing the inherent challenges of MCTS-based methods.



\begin{figure}[t]
    \centering
    \includegraphics[width=\linewidth]{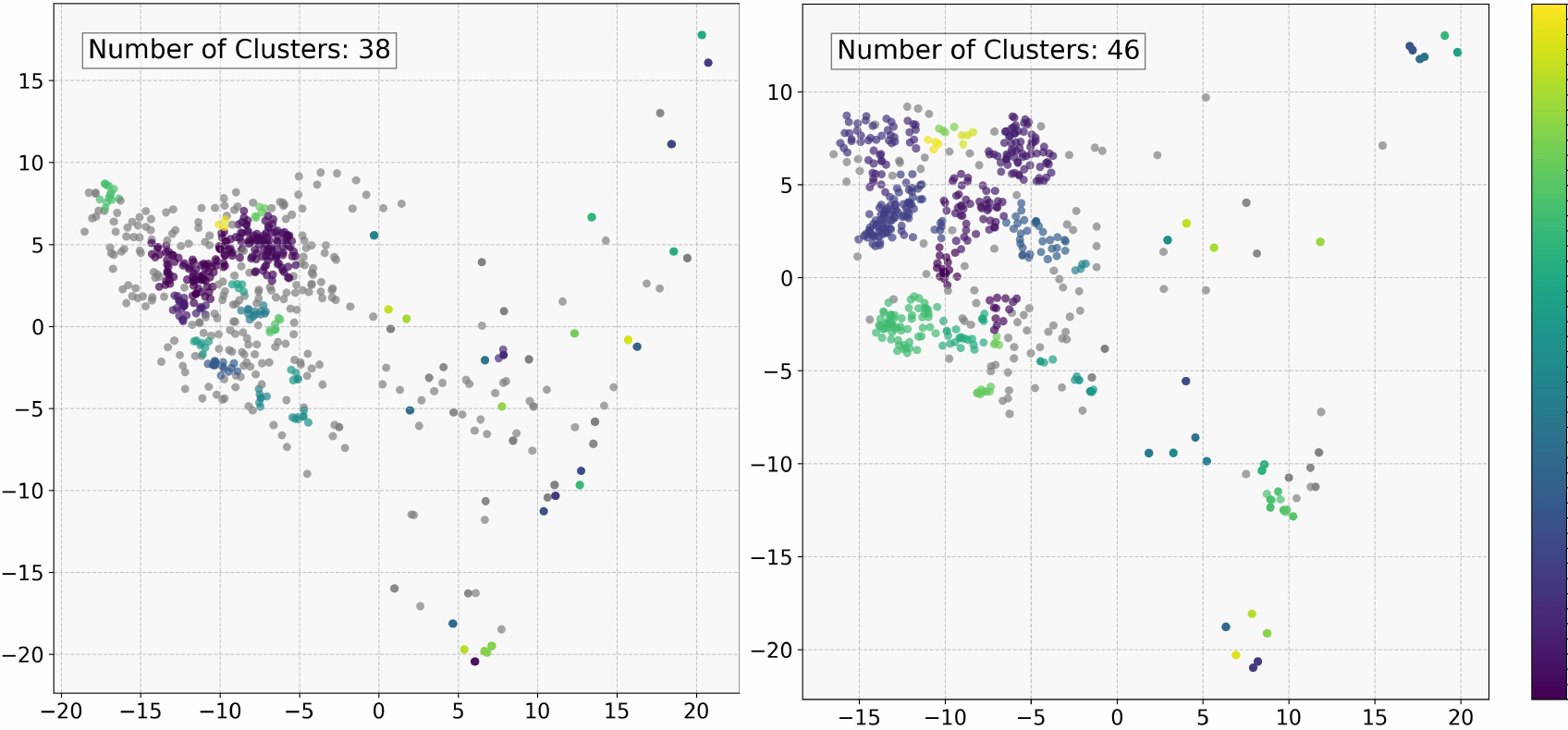}
     \vspace{-1.5em}
    \caption{The visualization of the cadidate reasoning paths.}
   
    \label{fig:visualize}
\end{figure}

\noindent\textbf{Diversity Analysis.}\quad To investigate whether AR-MCTS can truly enhance the diversity of sampled solutions, we sample 250 problems from \textsc{MathVista} and employ AR-MCTS to sample 4 candidate solutions for each problem, yielding a total of 1,000 samples. We employ BGE-M3~\citep{bge_m3} as our semantic embedding model and apply PCA for dimensionality reduction, followed by DBSCAN~\citep{dbscan} clustering for the visualization of all solution semantic representations.

Figure~\ref{fig:visualize} shows the visualization between the beam search (left) and AR-MCTS (right). The representations of candidate solutions sampled by beam search tend to collapse into a small area with several noise points (in gray), reflecting that the beam search may lead to redundancy in sampling. Under the same parameter settings, AC-MCTS clusters more centroids for the same problem set (38 vs.46) and exhibits a more dispersed representation distribution. This effectively demonstrates that AR-MCTS alleviates the issue of limited diversity in candidate solutions sampled, efficiently covering the problem-solving space and providing strong prior conditions for the simulation process of MCTS.

\section{Conclusion}
In this paper, we propose AR-MCTS, a universal framework dedicated to progressively improving the complex multimodal reasoning capabilities of MLLMs through active retrieval and Monte Carlo Tree Search. AR-MCTS leverages the MCTS algorithm alongside an active retrieval strategy, which automatically acquires high-quality step-wise reasoning annotations to progressively align a process reward model, ultimately enabling process-level multimodal reasoning verification. Experimental results demonstrate the effectiveness of AR-MCTS across various MLLMs and benchmarks, confirming its ability to optimize sampling diversity and verification accuracy, and providing a promising solution for reliable reasoning.

{
    \small
    \bibliographystyle{ieeenat_fullname}
    \bibliography{main}
}

\clearpage
\setcounter{page}{1}
\maketitlesupplementary

\setcounter{section}{0}
\renewcommand{\thesection}{\Alph{section}}

\tableofcontents

\section{More Details about AR-MCTS}
\label{more-ar-mcts}

\subsection{The Algorithm Workflow of AR-MCTS}

In this section, we will explore the overall workflow of AR-MCTS, highlighting its key components and steps involved in the retrieval and inference process. For each query \(q\), we begin by applying the ``Unified Retrieval'' module to extract key insights, denoted as \(D_\text{ins}\). These insights serve as a sub-corpus for performing active retrieval during the MCTS process. 

As outlined in Algorithm~\ref{retrieval}, distinct retrievers are utilized to handle text-only and multimodal corpora separately. The top-\(K\) knowledge retrieved from both routes is then combined to form \(D_\text{top-K}\). This set of documents undergoes further refinement through the ``Knowledge Concept Filter,'' yielding the final corpus of key insights, \(D_\text{ins}\).

Once the key insights are obtained, the AR-MCTS inference algorithm, detailed in Algorithm~\ref{inference}, is executed. During each expansion step \(t\), given a beam size \(B\), the top-\(B\) most relevant insights are retrieved from \(D_\text{ins}\). Each retrieved document is paired with the previous state \(s_{t-1}\) and fed into independent paths, where the multimodal large language model (MLLM), \(\mathcal{M}\), generates the next state. The Process Reward Model (PRM), \(\pi_\theta\), then evaluates the candidate states \(s_{(t, 1)}, \dots, s_{(t, B)}\) and assigns PRM scores to each. The state with the highest PRM score is appended to the selected path \(\mathcal{P}\). This process continues until a terminal state is reached, resulting in the final reasoning trajectory and the corresponding answer.

\begin{algorithm*}
    \caption{Unified Retrieval}
    \label{retrieval}
    \begin{algorithmic}[1]
    
    \Require{Query \(q\), hybrid-modal retrieval corpus \(D_\text{H}\), top-\(K\), cross-modal retriever \(R_c\), text-to-text retriever \(R_t\)}
    \Ensure{Top-K retrieved hybrid-modal samples \(D_\text{top-K}\)}
    \ForAll{\(d_i \in D_\text{H}\)} \Comment{Text-Only Retrieval}
        \State Query embedding \(E_q \gets R_t(q)\)
        \State Document embedding \(E_{d_i} \gets R_t(d_i)\)
        \State Retrieved documents \(D_\text{text} = \text{argtop}_K^{i = 1, \ldots, N} \left[ E_{d_i}^{\top} \cdot E_q \right]\)
    \EndFor
    \ForAll{image-text pair \((x,t) \in D_\text{H}\)} \Comment{Cross-modal Retrieval}
        \State Image embedding \(E_I(x) \gets R_c(x)\)
        \State Text embedding \(E_T(t) \gets R_c(t)\)
        \If{\(t \neq \emptyset \land x \neq \emptyset\)}
            \State \(E_x(x,t) \gets \frac{E_{I}(x) + E_{T}(t)}{2}\)
        \ElsIf{\(t \neq \emptyset \land x = \emptyset\)}
            \State \(E_x(x,t) \gets E_{T}(t)\)
        \EndIf
        \State Mixed vector \(E_x(Q^m) \gets \frac{E_{I}(x) + E_{T}(t)}{2}\)
        \State Retrieved documents \(D_{\text{cross}} \gets \text{argtop}_K^{j = 1, \ldots, N} \left[ E_{x}(Q^m)^{\top} \cdot E_{x}(x_j, t_j) \right]\)
    \EndFor
    \State \(D_\text{top-K} \gets \{D_{\text{q}} \cup D_{\text{cross}}\}\)
    \Require{Knowledge concept label \(L_\text{kc}\), original retrieval threshold \(T_r\), knowledge concept consistency threshold \(T_\text{kc}\)}
    \Ensure{Key insights \(D_{\text{ins}}\) for query \(q\)}
    \State Key insights \(D_{\text{ins}} = \{r \in D_{\text{H}} \mid \text{Sim}(r, q) \geq T_r \ \& \ \text{Sim}(r, L_{\text{kc}}) \geq T_{\text{kc}}\}\) \Comment{Knowledge Concept Filtering}
    
    \end{algorithmic}
\end{algorithm*}


\begin{algorithm*}
    \caption{Inference with AR-MCTS}
    \label{inference}
    \begin{algorithmic}[1]
    \Require Beam Size \(B\), question \(q\), Process Reward Model \(\pi_\theta\), max depth \(T\), MLLM \(\mathcal{M}\), multimodal retriever \(R\)
    \Ensure Selected path (thought process and answer) \(\mathcal{P}\)
    \State \(\mathcal{P} = [s_0], \, t = 0\) \Comment{Initialize Selected Path}
    \While{\(t < T \land \text{non-terminal path in } \mathcal{P}\)}
        \State Retrieved insights \(D_\text{ins}\)
        \State \(D_\text{top\_B} \gets R(\mathcal{P}, D_\text{ins})\) 
        \ForAll{\(d_i \in D_\text{top\_B}\)}
            \State \(s_{(t, i)} \gets \mathcal{M}(\mathcal{P}, d_i)\)
            \State PRM score \(\text{score}(s_{t, i}) = \pi_\theta(s_{t, i})\)
        \EndFor
        \State \(j \gets \text{index}(\text{argmax}(\text{score}))\)
        \State Add \(s_{(t, j)}\) to \(\mathcal{P}\)
        \State Increment \(t \gets t + 1\)
    \EndWhile
    \end{algorithmic}
\end{algorithm*}


\begin{table}[!t]
\centering
\renewcommand{\arraystretch}{1.2} 
\caption{The statistics of General Reasoning Knowledge.}
\label{tab:dataset_statistics1}

   \setlength{\tabcolsep}{5pt}{
\begin{tabular}{lcc}
  
\toprule
\textbf{Dataset}          & \textbf{Count} & \textbf{Percentage} \\ \hline
Wikipedia(zh-CN)          &  4.7B             & 23.9\%                   \\
Wikipedia(en-US)          & 15B             & 73.6\%                   \\
COIG                      & 178K           & 0.1\%                   \\
\bottomrule
\end{tabular}
}
\end{table}

\begin{table}[!t]
\centering



\caption{The statistics of Mathematics-Specific Reasoning Knowledge.}
     
\renewcommand{\arraystretch}{1.2}      
     \setlength{\tabcolsep}{5pt}{
\label{tab:dataset_statistics2}
\begin{tabular}{lcc}
\toprule

\textbf{Dataset}          & \textbf{Count} & \textbf{Percentage} \\ \hline

\multicolumn{1}{l}{\textit{Text-only Datasets}} \\
GSM8K                     & 8,792          & 24.6\%              \\
MATH                      & 12,500         & 36.2\%              \\ \hline

\multicolumn{1}{l}{\textit{Multimodal Datasets}} \\
\textsc{MathVista}                 & 6,141          & 17.8\%              \\
MathVerse                 & 2,612          & 7.6\%               \\
MathVision               & 3,040          & 8.8\%               \\
\textsc{We-Math}                   & 1,740          & 5.0\%               \\ 

\bottomrule
\end{tabular}
}
\end{table}

\section{More Details about Experimental Setup}
\label{more-setup}
\subsection{Benchmarks and Datasets}

Here are the details of the benchmarks/datasets we used in our hybrid-modal retrieved corpus and experiments. The statistics of the datasets are recorded in Table \ref{tab:dataset_statistics1} and \ref{tab:dataset_statistics2}.

\begin{itemize}
\item \textbf{\textsc{We-Math}}~\citep{WeMath} is a benchmark based on textbook knowledge units, focusing on decomposing complex problems into sub-problems using fundamental concepts. It mirrors how students learn progressively and is organized hierarchically, following textbook content to maintain independent knowledge units while establishing logical connections between levels. It uses diverse evaluation metrics to comprehensively assess models’ ability of solving multimodal mathematical problems step by step.

\item \textbf{\textsc{MathVista}}~\citep{MathVista} is a mathematical visual benchmark consisting of 6,141 examples. These examples are divided into two subsets: \textit{testmini} (1,000 examples), for which answers are provided, and \textit{test} (5,141 examples), for which answers are not publicly available. We use this dataset as a benchmark to evaluate visual understanding and compositional reasoning abilities. Additionally, we employ LLaVA-OneVision-70B to generate answers for the \textit{test} split, creating an in-domain corpus that can be used for retrieving answers in the \textit{testmini} set.

\item \textbf{MathVision}~\citep{MathVision} is a carefully curated dataset consisting of 3,040 high-quality mathematical problems, each accompanied by a visual context derived from real mathematics competitions. The collection covers 16 distinct mathematical domains and is categorized across five levels of difficulty. We use this dataset as part of math reasoning knowledge base.

\item \textbf{\textsc{MathVerse}}~\citep{MathVerse} is a comprehensive and specialized visual mathematics benchmark designed to evaluate the multimodal mathematical reasoning abilities of MLLMs. It comprises a dataset of 2,612 visual math problems, with 1,236 newly acquired from public question repositories and 1,376 sourced from existing benchmarks. Each problem has been transformed by human annotators into six distinct versions—text-dominant, text-lite, text-only, vision-intensive, vision-dominant, and vision-only—each offering different levels of multimodal information. In our study, we utilize the "vision-only" version as image data and the "text-only" version as textual data to construct a knowledge base. This dataset is employed solely for the purpose of knowledge base construction and not as a benchmark.

\item \textbf{MATH}~\citep{math} is a dataset comprising 12,500 challenging competition mathematics problems. Each problem includes a comprehensive step-by-step solution, which can be used to train models in generating answer derivations and explanations. The dataset features problems from various mathematics competitions, including the AMC 10, AMC 12, AIME, and others. We utilize this dataset as a text-only corpus for mathematical domain reasoning.

\item \textbf{GSM8K}~\citep{gsm8k} (Grade School Math 8K) is a dataset containing 8,500 high-quality, linguistically diverse grade school math word problems. This dataset was designed to support question-answering tasks for basic mathematical problems requiring multi-step reasoning. This dataset is also used as part of our text-only mathematics-specific reasoning corpus.

\item \textbf{COIG}~\citep{coig} (Chinese Open Instruction Generalist) is a set of Chinese instruction datasets to advance the training and fine-tuning of Chinese LLMs. COIG includes five key corpora: a manually verified translated instruction corpus (66,858 entries), an exam-based Chain-of-Thought (CoT) instruction corpus derived from national exams (63,532 entries), a human value alignment corpus reflecting general and region-specific cultural values (34,471 entries), a counterfactual correction multi-round chat corpus addressing hallucinations and inconsistencies (13,653 dialogues), and a Leetcode instruction corpus supporting code-related tasks (11,737 entries). We utilize this dataset along with the Wikipedia corpus (English version and Chinese version) as our general reasoning knowledge base.

\item \textbf{GAOKAO-MM}~\citep{gaokaomm} is a comprehensive Chinese multimodal benchmark designed based on the Chinese National College Entrance Examination (Gaokao). It encompasses eight academic subjects and includes twelve categories of images, such as diagrams, function graphs, maps, and photographs. The benchmark aims to evaluate models' abilities to understand and reason over diverse multimodal content, reflecting the complexity and breadth of knowledge. We construct the domain-specific knowledge base using questions and answers from the years 2010 to 2021, while employing the questions from 2022 and 2023 as the test set.

\end{itemize}

\subsection{Baselines and Backbone Models}
To assess the gains from our approach, we compare against a number of baselines as follows.

\begin{itemize}
\item \textbf{Self-Consistency}~\citep{self-consistency} involves sampling multiple reasoning paths from a large language model. Since each path may lead to different final answers, Self-Consistency selects the most consistent answer as the final output by marginalizing these sampled paths. This method is based on the intuition that complex reasoning problems often have a unique correct answer that can be reached through various approaches.

\item \textbf{Self-Correction}~\citep{Self-Refine} is an iterative refinement method that improves the output of large language models (LLMs) or Multimodal large language models (MLLMs) through self-feedback. The core idea is to mimic the human revision process in writing: first, a preliminary output is generated, feedback is provided on this output, and improvements are made iteratively based on the feedback. Notably, this process allows for iterative optimization.

\item \textbf{ORM}~\citep{ORM} samples data from the reasoning training set to obtain result-oriented annotations for each sampled path. These data are then used to train a verifier that assists the generator in identifying higher-quality reasoning paths during prediction. In this paper, we use the same data as for training PRM, with the distinction that the annotations are made directly using ground truth results rather than through AR-MCTS for step-level annotations, training ORM to assess the quality of reasoning paths.

\end{itemize}

In addition, we provide a detailed introduction to the MLLMs designed in our experiment and its corresponding foundational language model.

\begin{itemize}
\item \textbf{Qwen2-VL}~\citep{qwen2-vl}, developed by Alibaba Cloud, represents an advanced iteration of the Qwen-VL series. By employing the Naive Dynamic Resolution mechanism, it dynamically processes images with varying resolutions and aspect ratios. The model achieves state-of-the-art performance on several visual understanding benchmarks, including \textsc{MathVista}, MathVision, and \textsc{We-Math}.

\item \textbf{InternVL2}~\citep{internvl2} family comprises multimodal large language models designed for advanced multimodal understanding tasks, demonstrating performance competitive with proprietary systems. Built using a progressive alignment training strategy, InternVL2 supports multimodal inputs, generalizes across diverse downstream tasks, and spans models ranging in size from 1 billion to 108 billion parameters. The InternVL2-8B variant exhibits remarkable capabilities in complex reasoning and shows promise for mathematical problem-solving applications.

\item \textbf{LLaVA-NeXT}~\citep{liu2024llavanext} is a large-scale multimodal language model optimized through a cost-effective, realistic visual instruction-tuning dataset. It emphasizes enhanced visual reasoning, optical character recognition (OCR), and visual conversation capabilities. LLaVA-NeXT demonstrates superior performance across various multimodal benchmarks, including MMMU and \textsc{MathVista}.

\item \textbf{LLaVA-OneVision}~\citep{llava_onevision} is a family of large-scale multimodal large language models (MLLMs) designed to extend the performance boundaries of open MLLMs across diverse scenarios, including single-image, multi-image, and video applications. It processes text, images, interleaved image-text inputs, and videos, supporting resolutions of up to 2304×2304 pixels. LLaVA-OneVision is available in various sizes, ranging from 0.5 billion to 72 billion parameters, and facilitates robust task transfer learning across modalities. Notably, it demonstrates exceptional video understanding by leveraging task transfer capabilities developed from image-based training.

\item \textbf{GPT-4o}~\citep{GPT-4o}, a proprietary large-scale multimodal model developed by OpenAI, processes vision, text, and audio inputs. Built on Transformer architecture, the model is pre-trained on next-token prediction tasks and refined through a post-training alignment process. GPT-4o exhibits state-of-the-art multimodal understanding, achieving outstanding results across a variety of complex multimodal tasks.

\item \textbf{GPT-4V}~\citep{GPT-4V}, also developed by OpenAI, is a highly capable multimodal system enabling users to process image inputs and interleaved image-text data with GPT-4 models. It achieves impressive human-level performance across a broad spectrum of tasks, including scene text understanding, abstract reasoning, and open-world question answering.

\item \textbf{Qwen2}~\citep{Qwen2} is a series of large language models (LLMs) based on the Transformer architecture, trained on a high-quality, diverse dataset of over 7 trillion tokens using next-token prediction. Spanning parameter sizes from 0.5 billion to 72 billion, Qwen2 is designed to enhance mathematical and coding reasoning capabilities. It achieves performance competitive with proprietary models across benchmarks for reasoning, language understanding, and generation. The Qwen2 series includes both foundational models and instruction-tuned variants, fine-tuned on datasets for single-turn and multi-turn instruction following.

\item \textbf{InternLM2.5}~\citep{internlm2} is a series of LLMs optimized for superior mathematical reasoning. Based on the InternLM2.5 foundational models, the series includes chat models fine-tuned through supervised fine-tuning (SFT) and reinforcement learning from human feedback (RLHF), enabling robust instruction-following capabilities in downstream tasks. Notably, InternLM2.5-Chat-1M supports a 1-million-token context, demonstrating exceptional performance on long-context benchmarks.

\item \textbf{Llama3}~\citep{Llama-3} is a family of LLMs built on a standard dense Transformer architecture, natively supporting multilinguality, coding, reasoning, and tool integration. Pretrained on a large-scale, meticulously curated dataset, it undergoes post-training through supervised fine-tuning (SFT), rejection sampling (RS), and direct preference optimization (DPO). The flagship model, LLaMA3-405B, represents a significant scale-up from its predecessor LLaMA2, trained on 15.6 trillion text tokens. It delivers competitive performance with GPT-4 across diverse benchmarks, including GSM8k, MATH, and MMLU.

\end{itemize}

\subsection{Implementation Details}
For uni-modal retrieval, we utilize mcontriever-mscoco, a multilingual version of Contriever~\citep{contriever} fine-tuned on the MSMARCO dataset, as the text encoder. For multimodal retrieval, we employ the frozen CLIP model (ViT-L/14@336px variant)~\cite{CLIP} as the multimodal encoder for both texts and images. To ensure the diversity and relevance of multimodal retrieval results, we incorporate five types of similarity measures: text-to-text, text-to-image, image-to-image, image-to-text and cross-modal retrieval (introduced in Section 4.2) . Given the extensive size of the knowledge base, we leverage the open-source indexing engine FAISS~\citep{faiss-gpu} to efficiently index dense vectors and retrieve the Top-k knowledge pieces.

For the Curriculum Process Reward Modeling, in the "Step-Wise DPO Pre-Alignment" phase, the learning rate is set to 5e-7 with a cosine scheduler and a 0.1 warm-up ratio. We use DeepSpeed ZeRO Stage 3~\citep{deepspeed} and Flash-Attention 2~\citep{flashattention} for efficiency, with a global batch size of 64. Training utilizes a sigmoid loss function with a beta value of 0.3 and spans 2 epochs, with checkpoints every 500 steps. Mixed precision training with bf16 is employed, and the maximum context length is 4096 tokens.

In the "Point-Wise Fine-Tuning" phase, we perform full fine-tuning on our PRM with a learning rate of 7e-6, using a linear scheduler with 20 warm-up steps. All models are trained with DeepSpeed ZeRO Stage 3 and Flash-Attention 2. We use a global batch size of 128, a weight decay of 0.1, and train for 3 epochs, saving checkpoints every 200 steps. Mixed precision training with bf16 is used, and the maximum context length is set to 8192 tokens. We run all our experiments on 8 NVIDIA A800 GPUs.

\begin{figure*}[t]
    \centering
    \includegraphics[width=\textwidth]{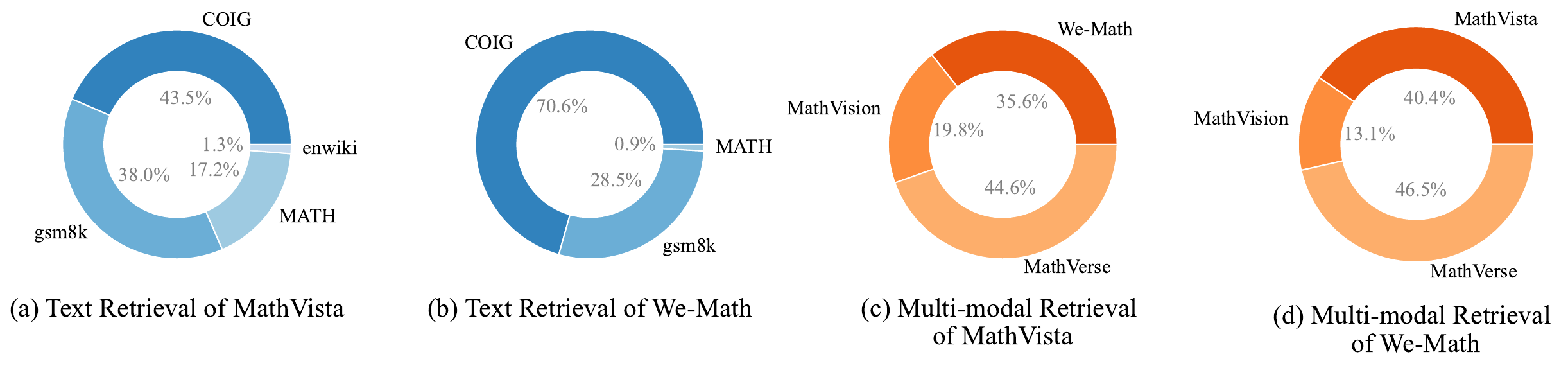}
    \vspace{-1.5em}
    \caption{The composition analysis on retrieval corpus of \textsc{We-Math} and \textsc{MathVista}.}
    \label{fig:kb}
    \vspace{-1em}
\end{figure*}

\subsection{Detailed Processing about Retrieved Corpus}

For Chinese evaluation on GAOKAO-MM, we utilize the COIG dataset and Chinese Wikipedia dump which contains over 2.6 million articles as the textual knowledge base. We first apply the tool WikiExtractor~\citep{wikiextractor} to extract clean texts from the Wikipedia dump and remove low-resource articles, which results in over 1.3 million articles. We then split each article into disjoint passages of 256 characters, resulting in 4.7B passages. To enrich our knowledge base with more relevant in-domain information, we split GAOKAO-MM questions from 2010 to 2021 as a multimodal knowledge base. Since the evaluation is conducted on a more up-to-date subset GAOKAO-MM from year 2022 to 2023, we can effectively mitigate the risk of data leakage. 

For English evaluation on \textsc{MathVista} and \textsc{We-Math}, we choose COIG, GSM8K, MATH, and the English Wikipedia dump as the textual knowledge base. Following the same pre-processing steps of Chinese Wikipedia dump, we obtain over 15B passages from the English Wikipedia dump as the basic retrieval unit. We have opted to employ MathVerse and MathVision as the multimodal knowledge base for their relevance to mathematical problem-solving and comprehension. Following MRAG-COT~\citep{mrag-cot}, we use responses from LLaVA-OneVision~\citep{llava_onevision} as pseudo-answer for the test set of \textsc{MathVista}. Due to the absence of an appropriate high-quality multimodal reasoning retrieval source or a training set with answer annotation, we incorporate the testmini set of \textsc{We-Math} into the knowledge base of \textsc{MathVista} and include the testmini set of \textsc{MathVista} into the knowledge base of \textsc{We-Math}.

\subsection{PRM Training Guideline} 

Our PRM leverages the corresponding text backbone for evaluating MLLMs and consistently uses Qwen2-7B for closed-source models. Due to the lack of step supervision in multimodal reasoning, we collect existing open-source text-only PRM dataset, such as AlphaMath~\citep{AlphaMath-Almost-Zero}, Math-Shepherd~\citep{math-shepherd}, and PRM800K~\citep{VerifyStep}. We first follow previous text-only works~\citep{math-shepherd,ReST-MCTS*} and perform preliminary fine-tuning alignment on our PRM backbone. Using the pre-aligned LLM, we apply the annotated data \(D_{\text{align}}\) from AR-MCTS and follow section "Curriculum Process Reward Modeling" to finalize the PRM. Consequently, we do not perform targeted fine-tuning on any MLLMs with in-domain data; instead, we focus on optimizing the PRM, significantly reducing computational resource consumption.

\subsection{Detailes about Knowledge Concept Filtering}

As stated in the main text, high-quality labels are available for test sets like \textsc{MathVista} and \textsc{We-Math}. However, not all external retrieval libraries or evaluation datasets have fine-grained concept labels (e.g., Wikipedia). To ensure the scalability of concept filtering, we use the open-world Tagger \textit{InsTag}~\citep{instag} for offline knowledge concept annotation and repeat the aforementioned process for consistency filtering.

Specifically, we select the TagLM-13b-v2.0 model~\footnote{\href{https://huggingface.co/OFA-Sys/TagLM-13b-v2.0}{https://huggingface.co/OFA-Sys/TagLM-13b-v2.0}}. For text-only data, we directly annotated using InsTag and concatenated all coarse and fine-grained labels. For multimodal data, we generated captions for the images using the corresponding evaluation MLLM backbone, referencing InternVL2~\citep{internvl} and Vista~\citep{vista}. We design the following caption generation template: "This is an image of a reasoning question; can you provide a detailed description of the image content?" We then concatenated the captions with the text and further used InsTag for annotation. After obtaining fine-grained labels, we followed the process outlined in the ''Knowledge Concept Filtering'' section for consistency screening.

\section{More Details about Experimental Results}
\label{more-results}

\begin{table}[t!] 
\centering 
\footnotesize
   \renewcommand{\arraystretch}{1.2} 
  \setlength{\tabcolsep}{3.5pt} 
    \caption{Mathematical evaluation on \textsc{MathVista} \textit{testmini} sets. We select 6 out of the original 12 mathematical categories in \textsc{MathVista}: ALL (overall accuracy), GPS (geometry problem solving), MWP (math word problems), ALG (algebraic reasoning), GEO (geometry reasoning), and STA (statistical reasoning). In the results for each model, the best accuracy scores are highlighted in \textbf{bold}.}
    \vspace{-1em}
        \begin{tabular}{llcccccc}
			\toprule
            Model & Method & ALL & GPS & MWP & ALG & GEO & STA \\
            
            \cmidrule{1-8}
            \multirow{5}*{\makecell[l]{GPT-\\4V}}
            & Zero-shot & 53.7 & 59.6 & 53.8 & 59.8 & 58.2 & 58.5 \\
            & Self-Consistency & 56.2 & 65.4 & 53.2 & 63.7 & \textbf{63.2} & 58.8 \\
            & Self-Correction & 50.4 & 56.3 &50.2  & 55.9 & 56.1 & 57.4  \\
            & ORM &56.6  & 65.3 & 53.1 & \textbf{65.2} & \textbf{63.2} & 59.0 \\
            & \cellcolor{gray!15}AR-MCTS & \cellcolor{gray!15}\textbf{57.4} & \cellcolor{gray!15}\textbf{66.1} & \cellcolor{gray!15}\textbf{53.9} & \cellcolor{gray!15}64.8 & \cellcolor{gray!15}\textbf{63.2} & \cellcolor{gray!15}\textbf{59.5} \\

            \midrule
            \multirow{5}*{\makecell[l]{LLaVA-\\NEXT}}
            & Zero-shot & 22.5 & 22.3 & 13.4 & 24.4 & 24.7 & 22.3 \\
            & Self-Consistency & 23.1 & 22.6 & 16.7 & 26.0 & 24.3 & 24.3 \\
            & Seld-Correction & 22.5 & 22.6 & 17.2 & 24.9 & 22.6 & 25.2  \\
            & ORM & 24.4 &22.6  & \textbf{17.5} & 27.9 &24.3  & 29.9 \\
            & \cellcolor{gray!15}AR-MCTS & \cellcolor{gray!15}\textbf{25.6} & \cellcolor{gray!15}\textbf{23.0} & \cellcolor{gray!15}17.4 & \cellcolor{gray!15}\textbf{28.1} & \cellcolor{gray!15}\textbf{28.6} & \cellcolor{gray!15}\textbf{31.5} \\
            
			\bottomrule
    \end{tabular}
    \label{tab:app_mathvista_results}
        \vspace{-1em}
\end{table}

\subsection{Results on More MLLMs backbones}

To further validate the scalability of AR-MCTS, we conduct generalization studies on the widely used open-source MLLM Llama3-Llava-Next-8B and the powerful closed-source MLLM GPT-4V using \textsc{MathVista}. As shown in Table \ref{tab:app_mathvista_results}, AR-MCTS continues to achieve stable improvements and aligns with the three core conclusions from our main experiments:
\begin{itemize}
\item 1. MLLMs struggle to self-correct reasoning errors.
\item 2. PRM outperforms ORM in complex reasoning tasks.
\item 3. AR-MCTS better unlocks the reasoning potential of weaker MLLMs.
\end{itemize}
This further confirms the scalability of our core experimental findings.

\begin{table}[t!]
    \centering
    \small
        \renewcommand{\arraystretch}{1.1} 
     \caption{The contamination analysis on hybrid-modal retrieval corpus.}
    \begin{tabular}{lcc}
        \toprule
        \textbf{Dataset} & \textsc{\textbf{MathVista}} & \textsc{\textbf{We-Math}} \\
        \midrule
        \multicolumn{3}{l}{\textit{Text-only Datasets}} \\
        \textsc{COIG} & 0.1\% & 0.1\%   \\
        \textsc{Wikipedia(en-US)} & 0.6\% & 1.1\%  \\
        \textsc{GSM8K} & 4.5\%  & 2.0\% \\
        \textsc{MATH} & 4.5\%  & 1.8\% \\
        \midrule
        \multicolumn{3}{l}{\textit{Multimodal Datasets}} \\
        \textsc{MathVerse} & 0.7\%  & 2.9\%  \\
        \textsc{MathVision} & 0.3\%  & 0.9\%  \\
        \textsc{We-Math} & 0.5\% & - \\
        \textsc{MathVista}\textit{-testmini} & - & 4.2\%  \\
        \bottomrule
    \end{tabular}
   
    \label{tab:contamination}
\end{table}

\subsection{The Composition Analysis of Retrieval Knowledge}

To gain deeper insights into which knowledge sources provide the greatest benefits to our multimodal reasoning test set, we conduct a comprehensive ranking of the hybrid-modal knowledge retrieved from samples of \textsc{MathVista} and \textsc{We-Math} based on similarity, selecting the Top-50 relevant knowledge samples and visualizing their respective sources. As shown in Figure \ref{fig:kb}, both \textsc{MathVista} and \textsc{We-Math} exhibit significant diversity in their retrieved knowledge, whether from text-only or multimodal sources. This highlights the motivation for constructing our hybrid-modal retrieval library from diverse, high-quality reasoning datasets. It also confirms that the insights needed for problem-solving do not necessarily originate from the same type of data source but should be enhanced through diverse reasoning knowledge. Our hybrid-modal reasoning retrieval library effectively addresses this need.

\subsection{Contamination Analysis on Hybrid-modal Retrieval Corpus}
To further ensure that our hybrid-modal retrieval corpus does not contain any data leakage examples from the test set, we perform a data contamination analysis. We employ commonly used n-gram contamination algorithms to assess the overlap between the Top-50 samples retrieved by the retriever from different data sources and various test sets. As shown in Table~\ref{tab:contamination}, we follow the AUTOIF~\citep{autoif} and test the n-gram threshold of 13. The results show that all data sources exhibited an overlap of less than 5\% with \textsc{MathVista} and \textsc{We-Math}. This highlights that there is no overlap between our retrieval library and the test sets.

\subsection{Ablations on Different Retrievers}

To validate the effectiveness of our general retrieval component, we conduct ablation studies by replacing different text and multimodal retrievers. Specifically, we used the following:

\begin{itemize}
    \item \textbf{Text Retrievers:} BM25 (Sparse), Contriever (Dense)  
    \item \textbf{Multimodal Retrievers:} CLIP-ViT-L/14, Jina-CLIP-v1~\citep{jinaclip}
\end{itemize}

The experimental results are presented in the Table \ref{tab:text_retriever_ablation} and \ref{tab:mm_retriever_ablation}, where we concatenated the top two retrieval results for each sample. The results indicate that different retrievers provide varying degrees of enhancement for downstream reasoning tasks. This not only demonstrates that our general retrieval module is plug-and-play but also highlights the rationale behind our mixed retrieval library.

\begin{table}[!t]
    \centering
    \small
    \renewcommand{\arraystretch}{1.1} 
    \caption{The ablations of different text retrievers.}
    \label{tab:model_statistics}
     \setlength{\tabcolsep}{5pt}{
    \begin{tabular}{lcccccc}
        \toprule
        Model & ALL & GPS & MWP & ALG & GEO & STA \\ 
        \midrule
        Qwen2-VL-7B & 58.8 & 45.5 & 60.5 & 45.5 & 47.9 & 70.8 \\
     
        + BM25 & 60.2 & 54.8 & 57.9 & 53.3 & 54.6 & 72.1 \\ 
        + Contriever & 59.9 & 53.9 & 58.5 & 53.3 & 54.1 & 72.4 \\ 
        \bottomrule
    \end{tabular}
\label{tab:text_retriever_ablation}
    }
\end{table}

\begin{table}[!t]
    \centering
    \small
        \renewcommand{\arraystretch}{1.1} 
    \caption{The ablations of different multimodal retrievers.}
    \label{tab:model_statistics}
    \setlength{\tabcolsep}{4.5pt} 
    \begin{tabular}{lcccccccc}
        \toprule
        Model & S1 & S2 & S3   \\ 
        \midrule
        Qwen2-VL-7B &53.4 &37.2 &33.9  \\ 

        + CLIP-ViT-L/14 &54.9 & 38.7&34.5 &  \\ 
        + Jina-CLIP-v1 &54.4 &36.9 &34.1 & \\ 
        \bottomrule
    \end{tabular}

    \label{tab:mm_retriever_ablation}
\end{table}

\subsection{Comparison of Different Training Objectives for PRMs}

\begin{table}[!ht]
    \centering
    \small
    \renewcommand{\arraystretch}{1.1} 
    \caption{The comparison of different training objectives for PRMs. }
    \label{tab:model_statistics}
     \setlength{\tabcolsep}{5pt}{
    \begin{tabular}{lcccccc}
        \toprule
        Model & ALL & GPS & MWP & ALG & GEO & STA \\ 
        \midrule
        PRM (Hard) & 62.9 & 63.3& 71.5& 59.4& 62.2& 71.0\\   
        PRM (Soft) & 64.1& 63.9& 72.6& 60.9& 63.6& 72.4 \\ 
        \bottomrule
    \end{tabular}
\label{tab:soft_hard}
    }
\end{table}

In this section, we explore the relationship between the training method of PRM and its multimodal reasoning capabilities. Following the setup of \citeauthor{math_prove}, we investigate the use of hard labels versus soft labels trained through a linear layer connected to a large model.

As shown in Table \ref{tab:soft_hard}, we find that PRM trained with soft labels performs better than that trained with hard labels on \textsc{MathVista}.

\section{Limitations and Future Work}
Despite our efforts to optimize the AR-MCTS process, several limitations and areas for improvement remain:

\begin{itemize}
\item \textbf{Computational Cost Optimization:} Annotating processes with MCTS algorithms requires significant computational resources, leading to high resource consumption—a common challenge in reasoning verification. However, AR-MCTS costs are still substantially lower than manual annotation. As a plug-and-play framework, AR-MCTS focuses on optimizing reasoning without the need to train multimodal foundational models, which significantly reduces computational overhead. The emergence of efficient techniques like vLLM~\citep{kwon2023efficient} is also helping to address this issue.

\item \textbf{Exploration of PRM for Multimodal Model Foundations}: AR-MCTS represents a pioneering effort in stepwise reasoning within the multimodal domain, utilizing foundational training of MLLMs to align the PRM process. An ideal scenario would involve training the PRM within these models to enhance interactions between image and text and provide supplemental information for stepwise reasoning. However, the lack of annotated data and higher computational demands present significant challenges in this area, which remains largely unexplored and is a direction for our future work.

\item \textbf{Deep Integration of Retrieval and Reasoning}: Research highlights knowledge gaps in stepwise reasoning~\citep{rag_help_reasoning}. AR-MCTS introduces a dynamic retrieval strategy that effectively addresses this issue. We believe this area still holds great potential for exploration, particularly in dynamically supplementing missing knowledge based on feedback from multimodal large models, which will be a key focus of our future research.

\end{itemize}

\end{document}